\def\year{2022}\relax
\documentclass[letterpaper]{article} 
\usepackage{aaai22}  
\usepackage{times}  
\usepackage{helvet}  
\usepackage{courier}  
\usepackage[hyphens]{url}  
\usepackage{graphicx} 
\usepackage{natbib}  
\usepackage{caption} 
\DeclareCaptionStyle{ruled}{labelfont=normalfont,labelsep=colon,strut=off} 
\usepackage{algorithm}
\usepackage{algorithmic}

\usepackage{amsmath, amssymb}
\usepackage{multirow, booktabs}
\usepackage{subcaption}

\newcommand{\V}{${\mathcal V}$}
\newcommand{\E}{${\mathcal E}$}

\usepackage{newfloat}
\usepackage{listings}
\floatstyle{ruled}
\newfloat{listing}{tb}{lst}{}
\floatname{listing}{Listing}
\title{Defending Graph Convolutional Networks against Dynamic Graph Perturbations via Bayesian Self-supervision}

\author{Jun Zhuang, Mohammad Al Hasan}
\affiliations{
    Indiana University-Purdue University Indianapolis, IN, USA \\
    junz@iu.edu, alhasan@iupui.edu
}

\begin{document}

\maketitle

\begin{abstract}
In recent years, plentiful evidence illustrates that Graph Convolutional Networks (GCNs) achieve extraordinary accomplishments on the node classification task. However, GCNs may be vulnerable to adversarial attacks on label-scarce dynamic graphs. Many existing works aim to strengthen the robustness of GCNs; for instance, adversarial training is used to shield GCNs against malicious perturbations. However, these works fail on dynamic graphs for which label scarcity is a pressing issue. To overcome label scarcity, self-training attempts to iteratively assign pseudo-labels to highly confident unlabeled nodes but such attempts may suffer serious degradation under dynamic graph perturbations.
In this paper, we generalize noisy supervision as a kind of self-supervised learning method and then propose a novel Bayesian self-supervision model, namely GraphSS\footnote{Source code: \textbf{\url{https://github.com/junzhuang-code/GraphSS}}}, to address the issue.
Extensive experiments demonstrate that GraphSS can not only affirmatively alert the perturbations on dynamic graphs but also effectively recover the prediction of a node classifier when the graph is under such perturbations. These two advantages prove to be generalized over three classic GCNs across five public graph datasets.
\end{abstract}

\section{Introduction}
\label{sec:intro}

Graph Convolutional Networks (GCNs) have been widely ascertained to achieve extraordinary accomplishments on various node classification tasks on large graphs~\cite{kipf2016semi, du2017topology, bruna2013spectral, defferrard2016convolutional, wu2019simplifying, hamilton2017inductive}. For example, GCN-based node classification can be used for clustering users into different groups based on their interests and behavior in online social networks (OSNs), such as Facebook and Twitter. To do so, unlabeled nodes are annotated so that a supervised node classification model can be built, which can subsequently be used for advertisement, product recommendation, etc. 
However, GCNs may be vulnerable to adversarial attacks on label-scarce dynamic graphs. The label-scarce issue arises in a dynamic graph as new nodes appear, which are not annotated; lacking annotated labels undermine the training of robust GCNs~\cite{hu2019strategies}. Besides, temporal changes of the graph structure make GCNs vulnerable to adversarial attacks~\cite{dai2018adversarial, du2018robust} on any nodes at any time. We denote such adversarial attacks as Dynamic Graph Perturbations (DGP). Our objective in this work is to safeguard GCNs' performance in presence of Dynamic Graph Perturbations.

Extensive researches have been pursued to improve the robustness of GCNs on node classification tasks \cite{sun2018adversarial}. Such efforts can be categorized into four categories: adversarial learning methods~\cite{jin2019latent, deng2019batch, feng2019graph,   wang2019adversarial, zhang2020defensevgae, liu2018adv}, self-supervised learning methods~\cite{you2020graph, hassani2020contrastive, sun2020multi}, topological denoising methods~\cite{rong2019dropedge, luo2021learning}, and mechanism designing methods~\cite{zhang2020gnnguard, geisler2020reliable, chen2021understanding, liu2021elastic, jin2021node}.
The adversarial learning methods improve the robustness of GCNs by training with adversarial samples. However, these methods can barely train a robust model under the circumstance of extensive label scarcity.
To overcome label scarcity, a kind of self-supervised learning method, namely pre-training~\cite{galke2019can, hu2019pre, hu2020gpt, qiu2020gcc, shang2019pre}, is introduced, which constructs a pretext task to help GCNs learn transferable graph representation through the pre-training stage. Nevertheless, it is still a challenge to design a generalized pretext task that can be beneficial to arbitrary downstream tasks. 
Topological denoising methods prune suspicious edges on the graph before feeding this graph into GCNs. Nonetheless, such methods are not efficient on dynamic graphs as the graph structure is always changing over time.
Mechanism designing methods mainly propose a message-passing mechanism, a.k.a. graph convolutional operator or node aggregator, to better classify the nodes. Such a mechanism sometimes highly depends on heuristics explorations.

Self-training~\cite{sun2019multi} is an extended form of pre-training. It assigns self-generated pseudo-labels to highly confident unlabeled nodes and then adds these nodes to the set of labeled nodes for the next training iteration. Nonetheless, accurate assignment of such pseudo-labels may suffer serious degradation when the graph is undergoing dynamic graph perturbations.
Learning graph representation with noisy labels is a potential solution to mitigate this issue~\cite{zhuang2020deperturbation, zhuang2022does}. In this paper, we assume that noisy labels include both manual-annotated labels and auto-generated labels. We argue that such noisy labels assigned to the vertices could be regarded as a kind of self-information for each node, which can help GCN to retain its performance in presence of DGP.
In fact, usage of noisy labels is a kind of defense, which is equivalent to recovering the original label distribution of a node on a perturbed graph, for which the node's label distribution has undergone a change due to perturbation. This intuition inspires us to propose a Bayesian self-supervision model, namely GraphSS, which recovers the original label distribution iteratively, to improve the robustness of a GCN-based node classifier against dynamic graph perturbations.
First, GraphSS utilizes self-information, i.e., noisy labels of the nodes, to build a robust node classifier (without knowing the ground-truth latent labels), helping the node classification in a label-scarce graph. Most importantly, GraphSS can recover the prediction of a node classifier by iterative label transition with auto-generated labels when the graph is undergoing dynamic graph perturbations.

We summarize the contributions of this paper as follows:
\begin{itemize}
  \item We generalize noisy supervision as a subset of self-supervised learning methods and propose a new Bayesian self-supervision model, namely GraphSS, to improve the robustness of the node classifier on the dynamic graph. To the best of our knowledge, our work is the first model that adapts Bayesian inference with self-supervision and significantly improves the robustness of GCNs against perturbations on label-scarce dynamic graphs.
  \item Extensive experiments demonstrate that GraphSS can 1) effectively recover the prediction of a node classifier against dynamic graph perturbations, and 2) affirmatively alert such perturbations on dynamic graphs. These two advantages prove to be generalized over three classic GCNs across five public graph datasets.
\end{itemize}

\section{Methodology}
\label{sec:method}
In this section, we introduce the methodology of our proposed model, Bayesian self-supervision model, GraphSS, as follows. We first introduce the notation and preliminary background and then theoretically analyze the Bayesian self-supervision. Besides, we explain our algorithm and analyze its time complexity.

\noindent
\textbf{\large Notations and Preliminaries.}
Given an undirected attributed graph $\mathcal{G}$ = $(\mathcal{V}$, $\mathcal{E})$, where \V\ = $\{ v_{1}, v_{2}, ..., v_{N} \}$ denotes the set of vertices, $N$ is the number of vertices in $\mathcal{G}$, and \E\ $\subseteq$ \V\ $\times$ \V\ denotes the set of edges between vertices. We denote $\mathbf{A} \in \mathbb{R}^{N \times N}$ as the symmetric adjacency matrix and $\mathbf{X} \in \mathbb{R}^{N \times d}$ as the feature matrix, where $d$ is the number of features for each vertex.
We define the label-scarce graph as an extreme case in which all ground-truth labels, hereby referred as {\bf latent labels} of the vertices $\mathcal{Z} \in \mathbb{R}^{N \times 1}$, are unobserved. 
We argue that manual annotation is a potential solution to this problem but human annotation unavoidably brings into noises \citep{misra2016seeing}. Another potential solution is to use a trained node classifier to label the vertices with pseudo-labels. This solution also yields noisy labels. We use $\mathcal{Y} \in \mathbb{R}^{N \times 1}$ to denote the {\bf noisy labels}, which include manual-annotated labels $\mathcal{Y}_m$ and auto-generated labels $\mathcal{Y}_a$.
Our task is to defend GCN-based node classification when its noisy labels (observed) deviate from its latent labels (unobserved). However, we assume that the entries of both $\mathcal{Y}$ and $\mathcal{Z}$ take values from the same closed category set. Below, we first discuss the variant of graph convolutional networks (GCNs) that we consider for our task.

The most representative GCN proposed by \citep{kipf2016semi} is our preferred GCNs' variant. The layer-wise propagation of this GCN is presented as follows:
\begin{equation}
\scriptsize
\mathbf{H}^{(l+1)} = \sigma \left( \mathbf{\tilde{D}}^{-\frac{1}{2}} \mathbf{\tilde{A}} \mathbf{\tilde{D}}^{-\frac{1}{2}} \mathbf{H}^{(l)} \mathbf{W}^{(l)} \right)
\label{eqn:gcn}
\end{equation}
In Equation (\ref{eqn:gcn}), $\mathbf{\tilde{A}} = \mathbf{A} + I_{N}$, $\mathbf{\tilde{D}} = \mathbf{D} + I_{N}$, where $I_{N}$ is the identity matrix and $\mathbf{D}_{i,i} = \sum_{j} \mathbf{A}_{i,j}$ is the diagonal degree matrix. $\mathbf{H}^{(l)} \in \mathbb{R}^{N \times d}$ is the nodes hidden representation in the $l$-th layer, where $\mathbf{H}^{(0)} = \mathbf{X}$. $\mathbf{W}^{(l)}$ is the weight matrix in the $l$-th layer. $\sigma(\cdot)$ denotes a non-linear activation function, such as ReLU.

In the K-class node classification task, we denote $\mathbf{z}_{n}$ as the latent label of node $v_{n}$ and $\mathbf{y}_{n}$ as the corresponding noisy label ($\mathbf{y}_{nk}$
is a one-hot vector representation of $\mathbf{y}_n$). In our setting, the GCN can be trained by the loss function $\mathcal{L}$:
\begin{equation}
\scriptsize
\mathcal{L} = - \frac{1}{N} \sum_{n=1}^{N} \mathcal{L}(\mathbf{y}_{n}, \textit{f}_{\theta}(v_{n})) = - \frac{1}{NK} \sum_{n=1}^{N} \sum_{k=1}^{K} \mathbf{y}_{nk} \ln \textit{f}_{\theta}(v_{nk})
\label{eqn:loss1}
\end{equation}
where $\textit{f}_{\theta}(\cdot) = softmax(\mathbf{H}^{(l)})$ is the prediction of the node classifier parameterized by $\theta$.

Our idea to build a robust GCN-based node classifier is to use label transition, in which a transition matrix of size $K\times K$ is learned, which reflects a mapping from $\mathcal{Y}$ to  $\mathcal{Z}$. We represent such a matrix by $\phi$ (the same term is also used to denote a label to label mapping function). Under the presence of $\phi$, the loss function $\mathcal{L}$ in Equation (\ref{eqn:loss1}) can be rewritten as follows:
\begin{equation}
\scriptsize
\mathcal{L} = - \frac{1}{N} \sum_{n=1}^{N} \mathcal{L}(\mathbf{y}_{n},  
\phi^{-1} \circ \textit{f}_{\theta}(v_{n}))
\label{eqn:loss2}
\end{equation}
However, learning accurate $\phi$ and $f_{\theta}$ is a difficult task as the latent labels of the nodes are not available. So, in our method, the $\phi$ 
and $f_{\theta}$ are iteratively updated to approximate the perfect one by using the Bayesian framework.

\begin{figure}[h]
  \centering
  \includegraphics[width=\linewidth]{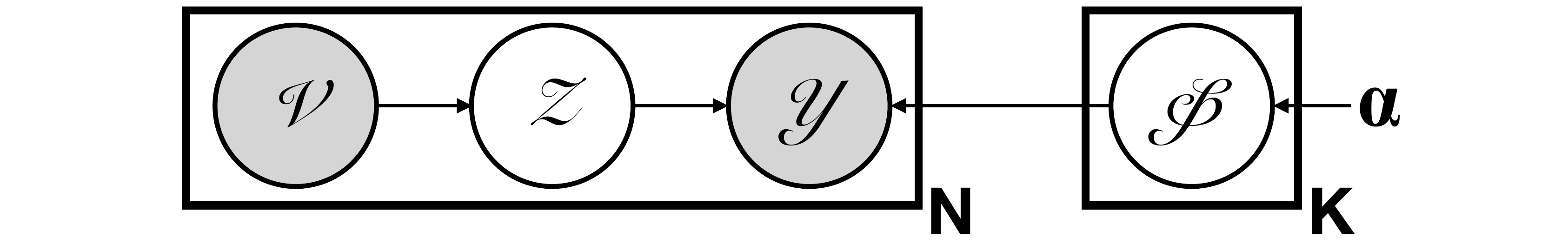}
  \caption{The diagram of Bayesian self-supervision ($\mathcal{V}$, $\mathcal{Z}$ and $\mathcal{Y}$ denote the nodes, the latent labels, and the noisy labels, respectively. $\phi$ denotes the conditional label transition matrix. $\alpha$ denotes the Dirichlet parameter. $N$ and $K$ denote the number of nodes and classes, respectively.)}
\label{fig:fig_ss}
\end{figure}

\noindent
\textbf{\large Bayesian Self-supervision.}
As shown in the plate diagram (Figure \ref{fig:fig_ss}), the unobserved latent labels ($\mathcal{Z}$) depend on the node features, whereas the observed noisy labels ($\mathcal{Y}$) depend on both $\mathcal{Z}$ and the conditional label transition matrix, $\phi$, modeled by $K$ multinomial distributions with a Dirichlet prior parameterized by $\alpha$.

The latent label of node $v_n$, $\mathbf{z}_{n} \sim \textit{P} \left( \cdot \mid v_{n} \right)$, where $\textit{P} \left(\cdot \mid v_{n} \right)$ is a $Categorical$ distribution modeled by the node classifier $\textit{f}_{\theta}(v_{n})$.
The noisy label $\mathbf{y}_{n} \sim \textit{P} \left( \cdot \mid \phi_{\mathbf{z}_{n}} \right)$, where $\phi_{\mathbf{z}_{n}}$ is the parameter of $Categorical$ distribution $\textit{P} \left( \cdot \mid \phi_{\mathbf{z}_{n}} \right)$.
The conditional label transition matrix $\phi$ = $[\phi_{1}, \phi_{2}, …, \phi_{K}]^{T}$ $\in \mathbb{R}^{K \times K}$ consists of K transition vectors. The $k$-th transition vector $\phi_{k} \sim Dirichlet(\alpha)$, where $\alpha$ is the parameter of $Dirichlet$ distribution.
The goal of our label transition is to obtain an updated $\textit{P} \left(\cdot \mid v_{n} \right)$ by using Bayesian self-supervision, so that the {\bf inferred label} of a given node sampled from this updated distribution is identical to the latent label of that node as much as possible.

According to the Figure \ref{fig:fig_ss}, we can employ Bayes' theorem to deduce the posterior of $\mathcal{Z}$ which is conditioned on the node $\mathcal{V}$, the noisy labels $\mathcal{Y}$, and the Dirichlet parameter $\alpha$.
\begin{equation}
\scriptsize
\begin{aligned}
\textit{P} \left( \mathcal{Z} \mid \mathcal{V}, \mathcal{Y} ; \alpha \right)
&= \textit{P} \left( \phi ; \alpha \right) \textit{P} \left( \mathcal{Z} \mid \mathcal{V}, \mathcal{Y}, \phi \right) \\
&= \int_{\phi}   \prod_{k=1}^{K} \textit{P} \left( \phi_{k} ; \alpha \right)   \prod_{n=1}^{N} \textit{P} \left( \mathbf{z}_{n} \mid v_{n}, \mathbf{y}_{n}, \phi \right) d\phi \\
&= \int_{\phi}   \prod_{k=1}^{K} \textit{P} \left( \phi_{k} ; \alpha \right)   \prod_{n=1}^{N} \frac{\textit{P} \left( \mathbf{z}_{n} \mid v_{n} \right) \textit{P} \left( \mathbf{y}_{n} \mid \mathbf{z}_{n}, \phi \right)}{\textit{P} \left( \mathbf{y}_{n} \mid v_{n} \right)} d\phi \\
\end{aligned}
\label{eqn:bayes1}
\end{equation}

According to the conjugation property between the Multinomial distribution and the Dirichlet distribution, we can continue to deduce the Equation (\ref{eqn:bayes1}) as follows:
\begin{equation}
\scriptsize
\begin{aligned}
&\textit{P} \left( \mathcal{Z} \mid \mathcal{V}, \mathcal{Y} ; \alpha \right) \\
&= \prod_{n=1}^{N} \frac{\textit{P} \left( \mathbf{z}_{n} \mid v_{n} \right)}{\textit{P} \left( \mathbf{y}_{n} \mid v_{n} \right)} \int_{\phi}   \prod_{k=1}^{K} \frac{\Gamma \left( \sum_{k'}^{K} \alpha_{k'} \right)}{\prod_{k'}^{K} \Gamma \left( \alpha_{k'} \right)}   \prod_{k'}^{K} \phi_{kk'}^{\alpha_{k'}-1}  \prod_{n=1}^{N} \phi_{\mathbf{z}_{n}\mathbf{y}_{n}}  d\phi \\
&= \prod_{n=1}^{N} \frac{\textit{P} \left( \mathbf{z}_{n} \mid v_{n} \right)}{\textit{P} \left( \mathbf{y}_{n} \mid v_{n} \right)} \int_{\phi}  \prod_{k=1}^{K} \frac{\Gamma \left( \sum_{k'}^{K} \alpha_{k'} \right)}{\prod_{k'}^{K} \Gamma \left( \alpha_{k'} \right)}   \prod_{k'}^{K} \phi_{kk'}^{\mathbf{C}_{kk'} + \alpha_{k'} - 1} d\phi \\
&= \prod_{n=1}^{N} \frac{\textit{P} \left( \mathbf{z}_{n} \mid v_{n} \right)}{\textit{P} \left( \mathbf{y}_{n} \mid v_{n} \right)}  \prod_{k=1}^{K} \frac{\Gamma \left( \sum_{k'}^{K} \alpha_{k'} \right)}{\prod_{k'}^{K} \Gamma \left( \alpha_{k'} \right)}   \prod_{k=1}^{K} \frac{\prod_{k'}^{K} \Gamma \left( \alpha_{k'} + \mathbf{C}_{kk'}  \right)}{\Gamma \left( \sum_{k'}^{K} \left( \alpha_{k'} + \mathbf{C}_{kk'} \right) \right)} \\
\end{aligned}
\label{eqn:bayes2}
\end{equation}
Here we denote the confusion matrix between the node prediction and the noisy labels as $\mathbf{C}$, where $\sum_{k}^{K} \sum_{k'}^{K} \mathbf{C}_{kk'} = N$. The term $ \prod_{n=1}^{N} \phi_{\mathbf{z}_{n}\mathbf{y}_{n}}$ is expressed as $ \prod_{k}^{K} \prod_{k'}^{K} \phi_{kk'}^{\mathbf{C}_{kk'}}$ so that we can integrate the terms based on the aforementioned conjugation property.

Unfortunately, Equation (\ref{eqn:bayes2}) can not directly be employed to infer the label. Instead, we apply Gibbs sampling here to approximate our goal.
According to Gibbs sampling, for each time we sample $\mathbf{z}_{n}$ by fixing $n$-th dimension in order to satisfy the detailed balance condition on the assumption of Markov chain \citep{bishop2006pattern}. Combined with Equation (\ref{eqn:bayes2}) and the recurrence relation of $\Gamma$ function, $\Gamma(n+1) = n\Gamma(n)$, we sample a sequence of $\mathbf{z}_{n}$ as follows:
\begin{equation}
\scriptsize
\begin{aligned}
\textit{P} \left( \mathbf{z}_{n} \mid \mathcal{Z}^{\neg \mathbf{z}_{n}}, \mathcal{V}, \mathcal{Y} ; \alpha \right) 
&= \frac{  \textit{P} \left( \mathcal{Z} \mid \mathcal{V}, \mathcal{Y} ; \alpha \right)  }{  \textit{P} \left( \mathcal{Z}^{\neg \mathbf{z}_{n}} \mid \mathcal{V}, \mathcal{Y} ; \alpha \right)  } \\
&= \frac{  \textit{P} \left( \mathbf{z}_{n} \mid v_{n} \right)  }{  \textit{P} \left( \mathbf{y}_{n} \mid v_{n} \right)  }
\frac{  \alpha_{\mathbf{y}_{n}} + \mathbf{C}_{\mathbf{z}_{n}\mathbf{y}_{n}}^{\neg \mathbf{z}_{n}}  }{  \sum_{k'}^{K} \left( \alpha_{k'} + \mathbf{C}_{\mathbf{z}_{n}k'}^{\neg \mathbf{z}_{n}} \right)  } \\
&\propto \overline{\textit{P}} \left( \mathbf{z}_{n} \mid v_{n} \right) \frac{  \alpha_{\mathbf{y}_{n}} + \mathbf{C}_{\mathbf{z}_{n}\mathbf{y}_{n}}^{\neg \mathbf{z}_{n}}  }{  \sum_{k'}^{K} \left( \alpha_{k'} + \mathbf{C}_{\mathbf{z}_{n}k'}^{\neg \mathbf{z}_{n}} \right)  } \\
\end{aligned}
\label{eqn:gibbs}
\end{equation}
where we denote $\mathcal{Z}^{\neg \mathbf{z}_{n}}$ as the subset of $\mathcal{Z}$ that removes statistic $\mathbf{z}_{n}$. In the last row of Equation (\ref{eqn:gibbs}), the first term $\overline{\textit{P}} \left(\mathbf{z}_{n} \mid v_{n}\right)$ is a categorical distribution of labels for the node $v_n$ modeled by $f_\theta$. We use the term $\overline{\textit{P}} \left(\mathcal{Z} \mid \mathcal{V} \right) \in \mathbb{R}^{N \times K}$ to denote the same over all the nodes. Note that $\textit{P} \left(\mathcal{Z} \mid \mathcal{V} \right) \in \mathbb{R}^{N \times 1}$ in this paper denotes the predicted labels, a.k.a. auto-generated labels $\mathcal{Y}_a$. To avoid confusion, we use the term $\mathcal{Y}_a$ to denote these labels in the rest of this paper. The second term represents the conditional label transition which is obtained from the posterior of the multinomial distribution corresponding to label transition from $\mathbf{y}_n$ to $\mathbf{z}_n$. We use Equation (\ref{eqn:gibbs}) to sample the inferred label, $\mathbf{z}_n$, which becomes the node $v_n$'s label for retraining $f_\theta$. Also, $\phi$ is updated through Bayesian inference in each iteration. Such process is repeated for a given number of epochs with the expectation that subsequent inferred label can approximate to the latent label.

\begin{figure}[h]
  \centering
  \includegraphics[width=\linewidth]{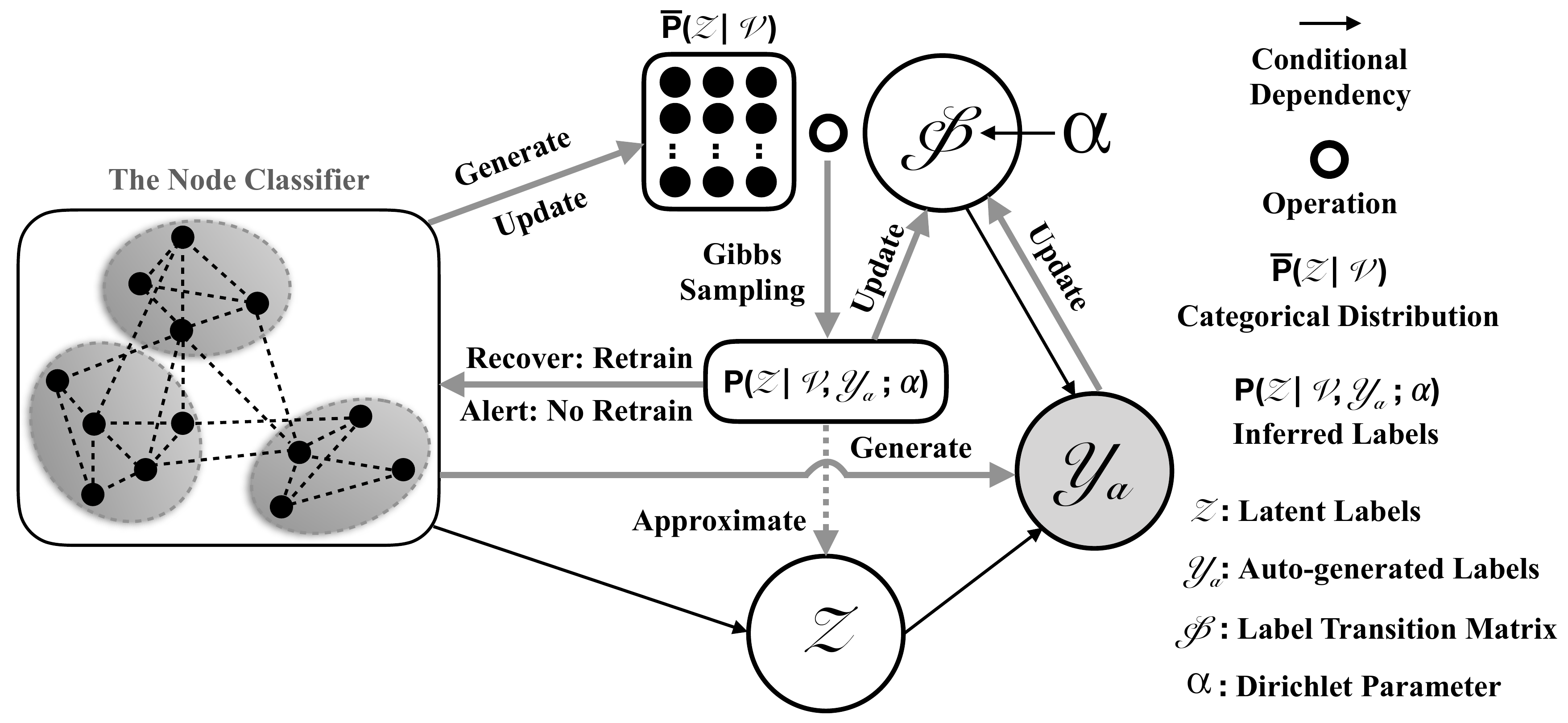}
  \caption{The workflow of Bayesian self-supervision model, GraphSS (The latent labels (white color) represent the unobserved ground-truth labels whereas the auto-generated labels (gray color) are observed noisy labels.)}
\label{fig:model}
\end{figure}

\noindent
\textbf{\large GraphSS Algorithm and Pseudo-code.}
The total process of GraphSS is displayed in Figure \ref{fig:model}. Given an undirected attribute graph, GraphSS classifies the nodes with manual-annotated labels $\mathcal{Y}_m$ at first and then generates both categorical distribution $\overline{\textit{P}} \left(\mathcal{Z} \mid \mathcal{V} \right)$ and auto-generated labels $\mathcal{Y}_a$. After that, GraphSS applies Gibbs sampling to sample the inferred labels $\textit{P} \left( \mathcal{Z} \mid \mathcal{V}, \mathcal{Y}_a ; \alpha \right) \in \mathbb{R}^{N \times 1}$ and updates the label transition matrix $\phi$ parameterized by $\alpha$. The information of \V\ is represented by both $\mathbf{A}$ and $\mathbf{X}$. For alert, GraphSS applies one-time label transition with $\mathcal{Y}_a$. For recovery, GraphSS iteratively re-trains the node classifier to update $\overline{\textit{P}} \left(\mathcal{Z} \mid \mathcal{V} \right)$. The inference will ultimately converge, approximating $\textit{P} \left( \mathcal{Z} \mid \mathcal{V}, \mathcal{Y}_a ; \alpha \right) $ to $\mathcal{Z}$ as close as possible. In brief, the goal of GraphSS is to sample the inferred labels by supervising the categorical distribution based on dynamic conditional label transition and ultimately approximates the inferred labels to the latent labels as close as possible.

\begin{algorithm}[tb]
\caption{{\sc Bayesian} Self-supervision}
\label{algo:graph_ss}
\textbf{Input}: Graph $\mathcal{G}_{train}$ and $\mathcal{G}_{test}$, which contain corresponding symmetric adjacency matrix $\mathbf{A}$ and feature matrix $\mathbf{X}$, Manual-annotated labels $\mathcal{Y}_m$ in $\mathcal{G}_{train}$, Node classifier $\textit{f}_{\theta}$, The number of warm-up steps $WS$, The number of epochs for inference $Epochs$
\begin{algorithmic}[1] 
\STATE Train $\textit{f}_{\theta}$ by Equation (\ref{eqn:loss1}) with $\mathcal{Y}_m$ on $\mathcal{G}_{train}$.
\STATE Generate categorical distribution $\overline{\textit{P}} \left(\mathcal{Z} \mid \mathcal{V} \right)$ and auto-generated labels $\mathcal{Y}_a$ by $\textit{f}_{\theta}$.
\STATE Compute the warm-up label transition matrix $\phi'$ based on $\mathcal{G}_{train}$.
\STATE Define the inferred labels $\textit{P} \left( \mathcal{Z} \mid \mathcal{V}, \mathcal{Y}_a ; \alpha \right) $ and the dynamic label transition matrix $\phi$ based on $\mathcal{G}_{test}$.
\FOR{$step \gets 1$ to $Epochs$}
  \IF {$step \leq WS$}
    \STATE {Sample $\mathbf{z}_{n}$ with warm-up $\phi'$ by Equation (\ref{eqn:gibbs}).}
  \ELSE
    \STATE {Sample $\mathbf{z}_{n}$ with dynamic $\phi$ by Equation (\ref{eqn:gibbs}).}
  \ENDIF
  \STATE {Update dynamic $\phi$, $\textit{P} \left( \mathcal{Z} \mid \mathcal{V}, \mathcal{Y}_a ; \alpha \right) $.}
  \IF {not Alert}
    \STATE {Retrain $\textit{f}_{\theta}$.}
  \ENDIF
\ENDFOR
\RETURN {$\textit{P} \left( \mathcal{Z} \mid \mathcal{V}, \mathcal{Y}_a ; \alpha \right)$ and Dynamic $\phi$.}
\end{algorithmic}
\end{algorithm}

The pseudo-code of GraphSS is shown in Algorithm (\ref{algo:graph_ss}).
{\bf Training:} GraphSS trains the node classifier $\textit{f}_{\theta}$ on the train graph $\mathcal{G}_{train}$ at first (\textbf{line 1}) with manual-annotated labels $\mathcal{Y}_m$ and then generates categorical distribution $\overline{\textit{P}} \left(\mathcal{Z} \mid \mathcal{V} \right)$ and auto-generated labels $\mathcal{Y}_a$ by $\textit{f}_{\theta}$ (\textbf{line 2}).
{\bf Inference:} Before the inference, GraphSS first computes a warm-up label transition matrix $\phi'$ by using the prediction over $\mathcal{G}_{train}$ (\textbf{line 3}) and then defines (creates empty spaces) the inferred labels $\textit{P} \left( \mathcal{Z} \mid \mathcal{V}, \mathcal{Y}_a ; \alpha \right) $ and the dynamic label transition matrix $\phi$ based on the test graph $\mathcal{G}_{test}$ (\textbf{line 4}). In the warm-up stage of the inference, GraphSS samples $\mathbf{z}_{n}$ with the warm-up $\phi'$ (\textbf{line 7}), which is built with the categorical distribution of $f_\theta$ and $\mathcal{Y}_m$ on $\mathcal{G}_{train}$. The categorical distributions of both $\mathcal{G}_{train}$ and $\mathcal{G}_{test}$ should have high similarity if both follow a similar distribution. Thus, the warm-up $\phi'$ is a key-stone since subsequent inference largely depends on this distribution. After the warm-up stage, GraphSS samples $\mathbf{z}_{n}$ with the dynamic $\phi$ (\textbf{line 9}). This dynamic $\phi$ updates in every epoch with current sampled $\mathbf{z}_{n}$ and corresponding $\mathcal{Y}_a$. Simultaneously, $\textit{P} \left( \mathcal{Z} \mid \mathcal{V}, \mathcal{Y}_a ; \alpha \right) $ is also updated based on the before-mentioned $\mathbf{z}_{n}$ (\textbf{line 11}). For recovery, GraphSS iteratively re-trains the node classifier to update its parameters (\textbf{line 13}). The inference will ultimately converge, approximating the inferred labels to the latent labels as close as possible. Note that both $\mathcal{G}_{train}$ and $\mathcal{G}_{test}$ contain corresponding symmetric adjacency matrix $\mathbf{A}$ and feature matrix $\mathbf{X}$. The categorical distributions of $\mathcal{G}_{test}$ may change abruptly when this graph is under perturbation since the original distribution in this graph is being perturbed with. In this case, GraphSS can also help recover the original categorical distribution by dynamic conditional label transition.

According to Algorithm (\ref{algo:graph_ss}), GraphSS applies Gibbs sampling via Equation (\ref{eqn:gibbs}) inside the $FOR$ loop. The time complexity of the sampling is $\mathcal{O} ( N_{test} \times K + K^{2} )$ since element-wise multiplication only traverses the number of elements in matrices once, where $N_{test}$ denotes the number of nodes in the test graph. In practice, the number of test nodes is far more than the number of classes, i.e., $N_{test} \gg K$. So, the time complexity of this sampling operation is approximately equal to $\mathcal{O}(N_{test})$. Hence, the {\bf time complexity} of inference except the first training (\textbf{line 1}) is $\mathcal{O}(Epochs \times N_{test})$, where $Epochs$ is the number of epochs for inference. 

\section{Experiments}
\label{sec:exp}
In this section, we present and analyze experimental results as follows. We first examine how far Bayesian self-supervision (GraphSS) can defend node classifiers against dynamic graph perturbations. We then assess whether GraphSS can alert such perturbations. Furthermore, we analyze the runtime, parameters and conduct an ablation study. At last, we discuss the limitation and future directions to illustrate how the community benefits from our work.

\begin{table}[h] 
\footnotesize
\centering
\begin{tabular}{cccccc}
  \toprule
    \textbf{Dataset} & {$\left| \mathcal{V} \right|$} & {$\left| \mathcal{E} \right|$} & {$\left| F \right|$} & {$\left| C \right|$} & {Avg.D} \\
    \midrule
    \textbf{Cora} & 2,708 & 10,556 & 1,433 & 7 & 3.85 \\
    \textbf{Citeseer} & 3,327 & 9,228 & 3,703 & 6 & 2.78 \\
    \textbf{Pubmed} & 19,717 & 88,651 & 500 & 3 & 4.49 \\
    \textbf{AMZcobuy} & 7,650 & 287,326 & 745 & 8 & 32.32 \\
    \textbf{Coauthor} & 18,333 & 327,576 & 6,805 & 15 & 10.01 \\
  \bottomrule
\end{tabular}
\caption{Statistics of datasets ($\left| \mathcal{V} \right|$, $\left| \mathcal{E} \right|$, $\left| F \right|$, and $\left| C \right|$ denote the number of nodes, edges, features, and classes, respectively. Avg.D denotes the average degree of test nodes.)}
\label{table:dataset}
\end{table}

\begin{table*}[t] 
\footnotesize
\centering
\begin{tabular}{c|c|c|c|c|c|c} 
\toprule %
 & \textbf{Test Acc. (\%)} & \textbf{Cora} & \textbf{Citeseer} & \textbf{Pubmed} & \textbf{AMZcobuy} & \textbf{Coauthor} \\
\midrule
\multirow{8}{*}{\textbf{GCN}} & Original & 81.46 (±0.89) & 69.95 (±0.53) & 81.83 (±0.62) & 93.14 (±0.96) & 92.01 (±0.85) \\
 & Attack~(Zügner et al. 2018) & 17.01 (±5.11) & 35.42 (±1.67) & 41.49 (±1.99) & 46.71 (±1.95) & 42.85 (±1.11) \\
 & AdvTrain~\citep{you2020does} & 25.64 (±2.16) & 37.66 (±1.80) & 44.06 (±0.79) & 52.50 (±1.40) & 62.18 (±0.55) \\
 & GNN-Jaccard~\citep{wu2019adversarial} & 62.31 (±2.14) & 68.79 (±1.66) & 69.73 (±1.24) & 82.76 (±2.38) & 79.42 (±0.57) \\
 & GNN-SVD~\citep{entezari2020all} & 54.65 (±1.82) & 63.92 (±2.91) & 63.75 (±1.10) & 78.99 (±1.48) & 73.64 (±0.99) \\
 & RGCN~\cite{zhu2019robust} & 31.72 (±2.03) & 46.93 (±1.60) & 48.19 (±2.32) & 50.90 (±1.84) & 57.51 (±1.22) \\
 & GRAND~\cite{feng2020graph} & 34.78 (±3.12) & 63.68 (±1.75) & 52.67 (±1.66) & 57.56 (±1.65) & 68.11 (±3.35) \\
 & ProGNN~\cite{jin2020graph} & 58.72 (±2.22) & 67.74 (±3.90) & 69.02 (±1.95) & 83.71 (±0.67) & 81.99 (±2.57) \\
 & NRGNN~(Dai et al. 2021) & 57.15 (±1.93) & 62.53 (±1.44) & 64.20 (±2.35) & 69.18 (±2.01) & 70.05 (±2.41) \\
 & GraphSS & \textbf{65.38 (±2.59)} & \textbf{69.18 (±0.99)} & \textbf{81.16 (±1.41)} & \textbf{90.88 (±1.61)} & \textbf{89.51 (±1.09)} \\
\midrule
\multirow{8}{*}{\textbf{SGC}} & Original & 83.21 (±1.52) & 69.65 (±0.86) & 83.43 (±1.12) & 90.56 (±1.25) & 91.04 (±1.31) \\
 & Attack~(Zügner et al. 2018) & 36.55 (±8.16) & 44.79 (±2.75) & 42.60 (±1.95) & 45.34 (±2.02) & 41.59 (±1.14) \\
 & AdvTrain~\citep{you2020does} & 46.28 (±3.84) & 47.43 (±2.58) & 53.65 (±1.69) & 50.42 (±1.30) & 62.52 (±1.82) \\
 & GNN-Jaccard~\citep{wu2019adversarial} & 84.35 (±3.74) & 69.77 (±1.59) & 72.48 (±1.28) & 79.72 (±2.22) & 79.61 (±1.16) \\
 & GNN-SVD~\citep{entezari2020all} & 65.42 (±4.20) & 54.39 (±1.64) & 68.40 (±1.70) & 75.65 (±2.16) & 72.54 (±1.24) \\
 & RGCN~\cite{zhu2019robust} & 41.32 (±0.67) & 45.17 (±3.07) & 46.56 (±3.22) & 48.42 (±1.68) & 56.33 (±1.29) \\
 & GRAND~\cite{feng2020graph} & 32.64 (±3.57) & 67.18 (±3.44) & 67.34 (±1.09) & 56.87 (±1.75) & 68.39 (±1.97) \\
 & ProGNN~\cite{jin2020graph} & 75.92 (±4.62) & 68.55 (±1.68) & 69.84 (±1.74) & 82.26 (±0.83) & 81.57 (±2.05) \\
 & NRGNN~(Dai et al. 2021) & 72.15 (±3.51) & 66.37 (±3.23) & 68.70 (±2.04) & 79.02 (±2.91) & 79.16 (±2.74) \\
 & GraphSS & \textbf{88.51 (±2.68)} & \textbf{70.78 (±3.17)} & \textbf{86.06 (±1.29)} & \textbf{88.70 (±1.39)} & \textbf{88.47 (±0.91)} \\
\midrule
\multirow{8}{*}{\textbf{GraphSAGE}} & Original & 84.50 (±1.61) & 72.58 (±0.94) & 84.99 (±1.08) & 91.99 (±0.96) & 91.98 (±1.02) \\
 & Attack~(Zügner et al. 2018) & 36.48 (±7.72) & 42.09 (±3.94) & 44.63 (±2.19) & 45.87 (±2.20) & 42.70 (±1.10) \\
 & AdvTrain~\citep{you2020does} & 47.78 (±7.58) & 42.66 (±1.77) & 53.54 (±1.99) & 51.60 (±1.09) & 62.70 (±1.83) \\
 & GNN-Jaccard~\citep{wu2019adversarial} & 84.97 (±3.85) & 69.99 (±1.96) & 74.52 (±1.25) & 81.81 (±2.06) & 78.72 (±0.71) \\
 & GNN-SVD~\citep{entezari2020all} & 66.03 (±4.19) & 57.92 (±0.42) & 69.87 (±1.33) & 77.42 (±1.63) & 72.25 (±1.47) \\
 & RGCN~\cite{zhu2019robust} & 41.98 (±0.69) & 45.08 (±1.67) & 47.04 (±2.50) & 50.29 (±1.95) & 57.23 (±1.43) \\
 & GRAND~\cite{feng2020graph} & 32.38 (±3.35) & 65.07 (±3.18) & 68.25 (±1.13) & 57.48 (±1.50) & 67.54 (±2.34) \\
 & ProGNN~\cite{jin2020graph} & 75.84 (±4.60) & 68.29 (±2.02) & 71.59 (±1.40) & 82.53 (±0.96) & 81.36 (±2.80) \\
 & NRGNN~(Dai et al. 2021) & 73.02 (±4.26) & 65.76 (±2.53) & 69.08 (±1.85) & 80.23 (±2.17) & 78.54 (±2.66) \\
 & GraphSS & \textbf{89.29 (±2.77)} & \textbf{73.43 (±2.32)} & \textbf{87.55 (±1.16)} & \textbf{90.20 (±1.79)} & \textbf{88.75 (±0.85)} \\
\bottomrule
\end{tabular}
\caption{The comparison of defending results (test accuracy) between GraphSS and the competing methods (Original/Attack denote the test accuracy before/after the non-target attacks ($L\&F$).)}
\label{table:exp1}
\end{table*}

\noindent
\textbf{\large Experimental Settings.}
Our proposed model is evaluated on five public datasets in Table \ref{table:dataset}. \textbf{Cora}, \textbf{Citeseer}, and \textbf{Pubmed} are famous citation graph data \citep{sen2008collective}. \textbf{AMZcobuy} comes from the photo segment of the Amazon co-purchase graph \citep{shchur2018pitfalls}. \textbf{Coauthor} is co-authorship graphs of computer science based on the Microsoft Academic Graph from the KDD Cup 2016 challenge \footnote{https://www.kdd.org/kdd-cup/view/kdd-cup-2016}.
For all five datasets, the percentage of train, validation, and test partition comprise 40\%, 20\%, and 40\% of the nodes, respectively. We select such partitions to simulate the OSNs as the size of the existing graph in OSNs will not be too small compared to the new coming dynamic subgraphs.
In this paper, we assume that the train graphs contain manual-annotated labels but the validation and test graphs don't. In other words, the node classifiers are trained with manual-annotated labels whereas GraphSS applies inference with auto-generated labels. To simulate the manual-annotated labels, we randomly replace the ground-truth label of a node with another label, chosen uniformly. We denote the percentage of such replacement as noise ratio $nr$ and fix $nr=10\%$ in the experiments. We set the number of training epochs as 200 because all node classifiers converge within 200 epochs in the training phase.
To simulate the dynamic graph perturbations, we execute node-level direct evasion non-targeted attacks on the links and features of the nodes ($L\&F$) in dynamic graphs \citep{zugner2018adversarial}, whereas the trained node classifier remains unchanged.

We compare our proposed model against six state-of-the-art defending methods.
AdvTrain~\cite{wang2019graphdefense, you2020does} assigns pseudo labels to generated adversarial samples and then retrains the node classifier with both noisy labeled nodes and adversarial nodes. 
GNN-Jaccard~\cite{wu2019adversarial} preprocesses the graph by eliminating suspicious connections, whose Jaccard similarity of node’s features is smaller than a given threshold.
GNN-SVD~\cite{entezari2020all} proposes another preprocessing approach with low-rank approximation on the perturbed graph to mitigate the negative effects from high-rank attacks, such as Nettack~\citep{zugner2018adversarial}.
RGCN~\cite{zhu2019robust} adopts Gaussian distributions as the hidden representations of nodes to mitigate the negative effects of adversarial attacks. 
GRAND~\cite{feng2020graph} proposes random propagation and consistency regularization strategies to address the issues about over-smoothing and non-robustness of GCNs. 
ProGNN~\cite{jin2020graph} jointly learns the structural graph properties and iteratively reconstructs the clean graph to reduce the effects of adversarial structure. 
NRGNN~\cite{dai2021nrgnn} develops a label noise-resistant framework that brings clean label information to unlabeled nodes based on the feature similarity.
Due to limited pages, we present implementation details, such as the hardware environment, the hyper-parameters of both our model and competing methods in supplementary materials.

\begin{figure*}[t]  
  \hfill
  \begin{subfigure}{0.195\textwidth}
  \centering 
    \includegraphics[width=\linewidth]{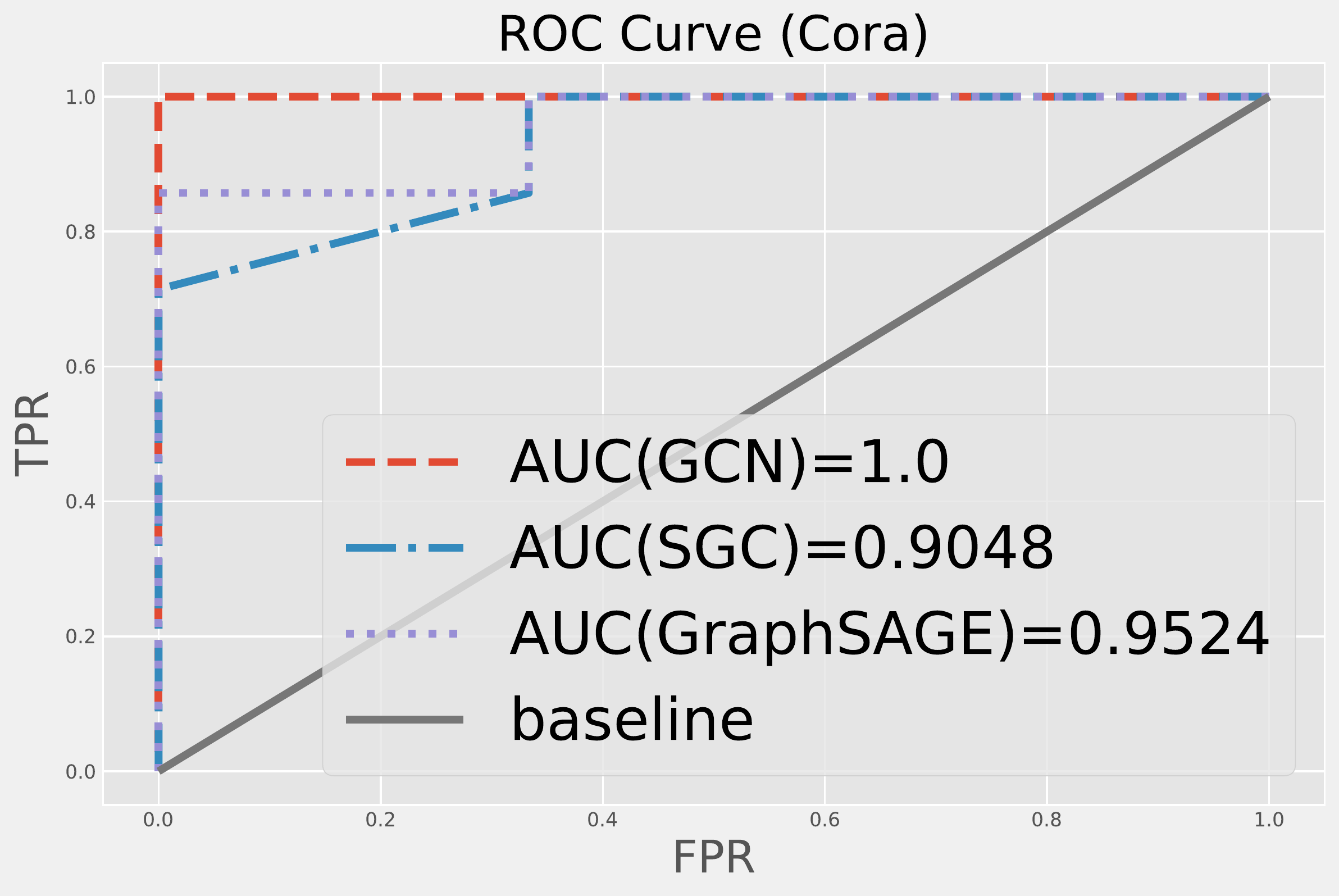}
  \end{subfigure}%
  \hfill
  \begin{subfigure}{0.195\textwidth}
  \centering 
    \includegraphics[width=\linewidth]{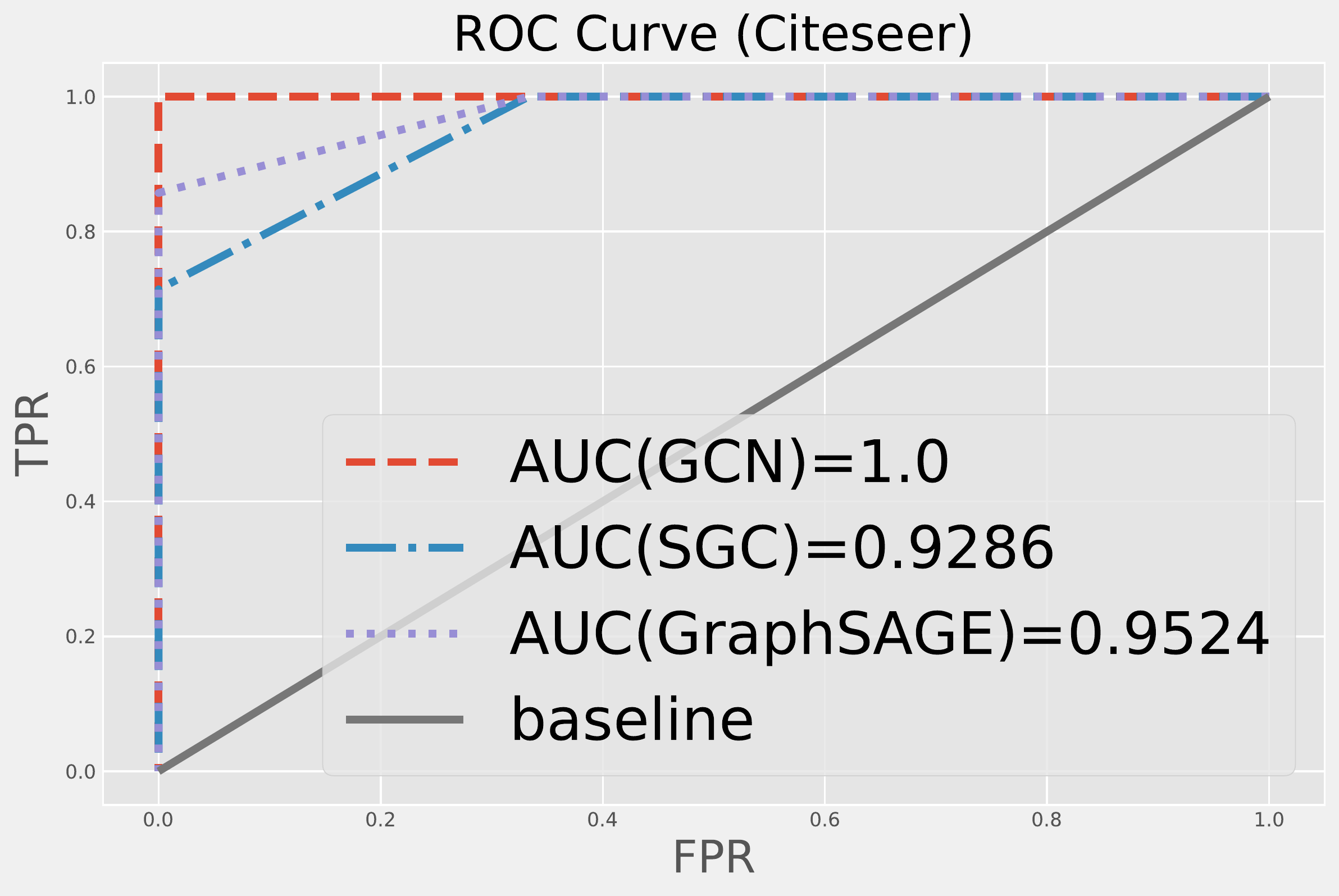}
  \end{subfigure}
  \begin{subfigure}{0.195\textwidth}
  \centering 
    \includegraphics[width=\linewidth]{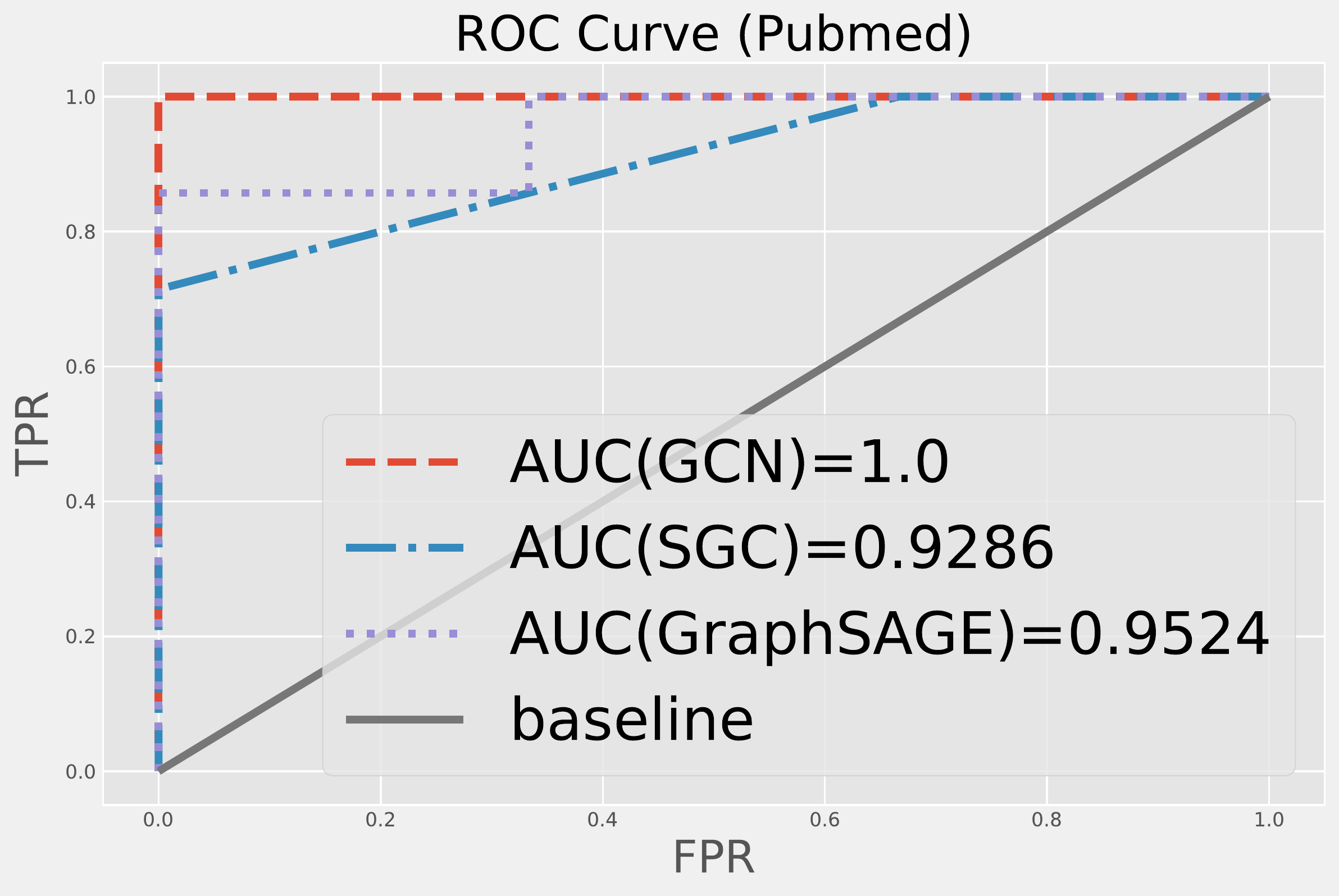}
  \end{subfigure}
  \begin{subfigure}{0.195\textwidth}
  \centering 
    \includegraphics[width=\linewidth]{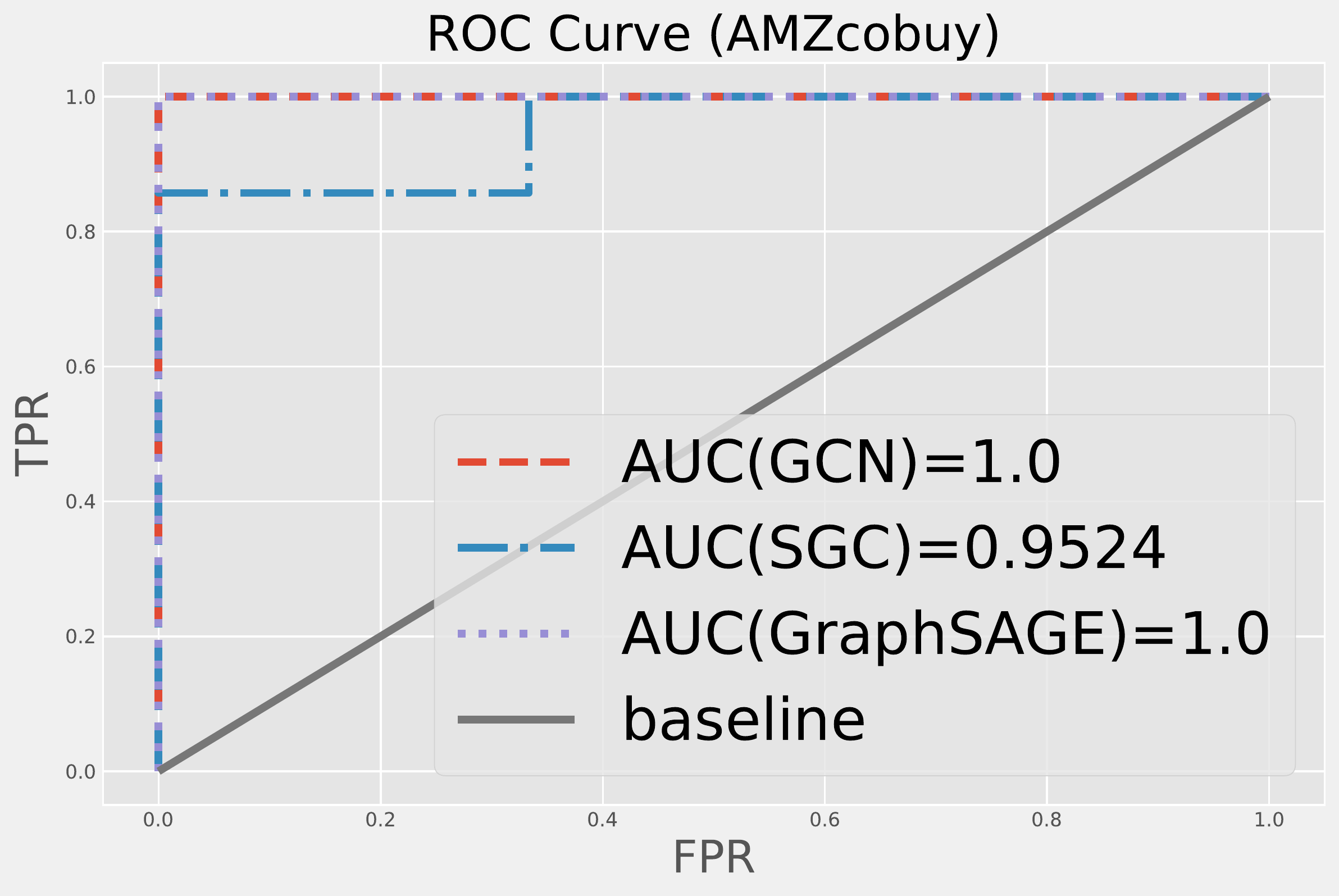}
  \end{subfigure}
  \begin{subfigure}{0.195\textwidth}
  \centering 
    \includegraphics[width=\linewidth]{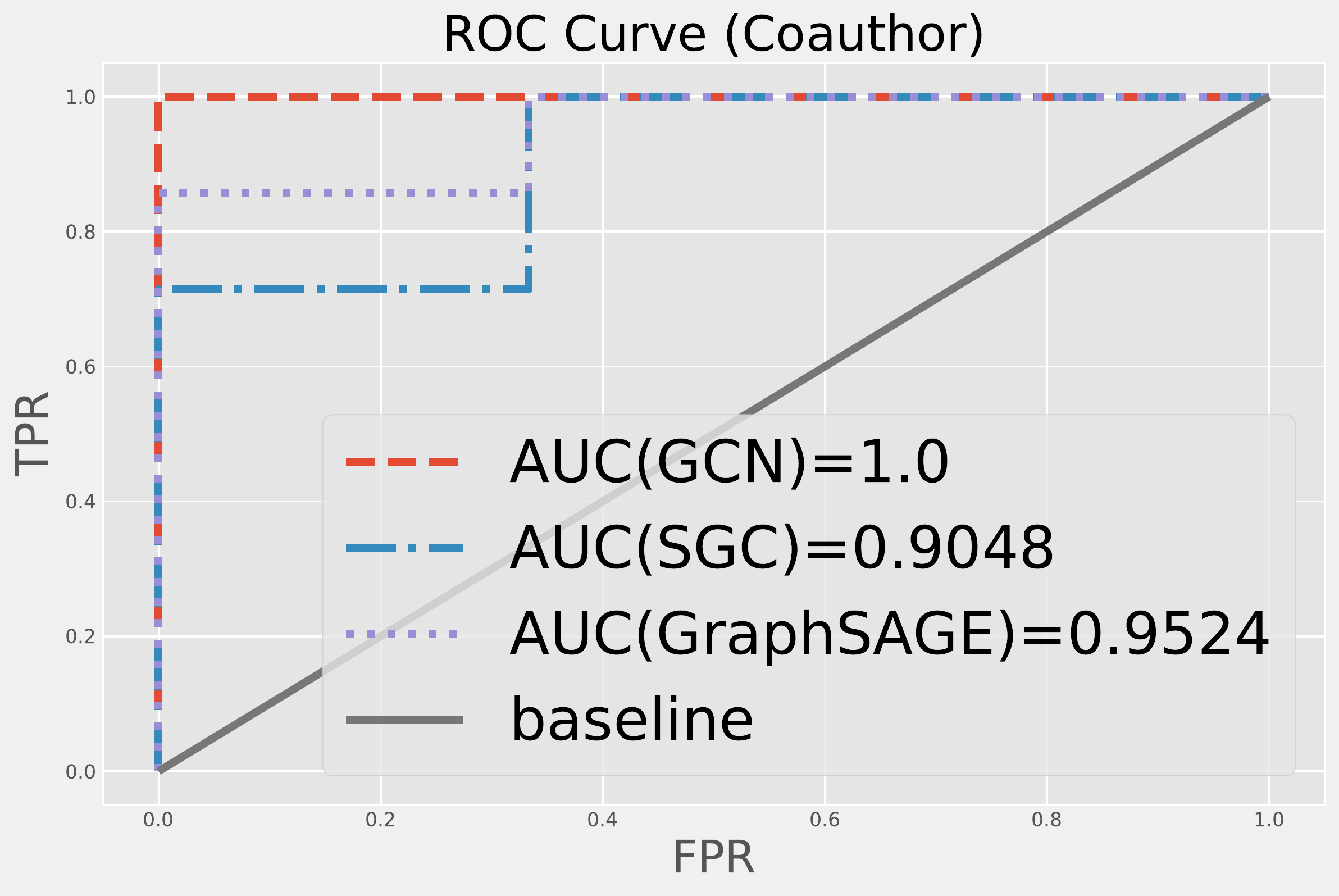}
  \end{subfigure}
\caption{The assessment of alerting dynamic graph perturbations by GraphSS}
\label{fig:fig_roc}
\end{figure*}

\begin{table*}[t]
\footnotesize
\centering
\setlength{\tabcolsep}{3pt}
\begin{tabular}{c|c|cc|cc|cc|cc|cc} 
\toprule %
\multirow{2}{*}{} & \multirow{2}{*}{\parbox{1.3cm}{\centering \textbf{Runtime (s)}}} & 
  \multicolumn{2}{c|}{\textbf{Cora}} &
  \multicolumn{2}{c|}{\textbf{Citeseer}} &
  \multicolumn{2}{c|}{\textbf{Pubmed}} &
  \multicolumn{2}{c|}{\textbf{AMZcobuy}} &
  \multicolumn{2}{c}{\textbf{Coauthor}} \\
\cline{3-12}
  &  & \textbf{Average} & \textbf{Unit} & \textbf{Average} & \textbf{Unit} & \textbf{Average} & \textbf{Unit} & \textbf{Average} & \textbf{Unit} & \textbf{Average} & \textbf{Unit} \\
\midrule
\multirow{2}{*}{\textbf{GCN}} & Defense & 22.75 (±0.65) & 0.84 & 27.86 (±0.66) & 0.84 & 163.57 (±3.29) & 0.83 & 61.19 (±0.66) & 0.80 & 146.90 (±1.07) & 0.80 \\
 & Alert & 21.28 (±0.31) & 0.79 & 26.14 (±0.44) & 0.79 & 155.84 (±1.57) & 0.79 & 61.05 (±0.60) & 0.80 & 146.27 (±1.83) & 0.80 \\
\midrule
\multirow{2}{*}{\textbf{SGC}} & Defense & 22.07 (±1.13) & 0.81 & 28.06 (±0.45) & 0.84 & 163.29 (±3.28) & 0.83 & 61.51 (±0.55) & 0.80 & 146.71 (±0.84) & 0.80 \\
 & Alert & 22.13 (±2.17) & 0.82 & 26.20 (±0.54) & 0.79 & 156.06 (±1.51) & 0.79 & 61.01 (±0.61) & 0.80 & 145.36 (±2.10) & 0.79 \\
\midrule
\multirow{2}{*}{\textbf{GraphSAGE}} & Defense & 22.04 (±0.91) & 0.81 & 28.51 (±0.59) & 0.86 & 165.44 (±0.86) & 0.84 & 61.75 (±0.45) & 0.81 & 147.28 (±0.77) & 0.80 \\
 & Alert & 21.38 (±0.22) & 0.79 & 26.06 (±0.36) & 0.78 & 154.94 (±1.04) & 0.79 & 61.12 (±0.70) & 0.80 & 147.14 (±1.36) & 0.80 \\
\bottomrule
\end{tabular}
\caption{Analysis of the average runtime and the unit runtime (per 100 nodes) of GraphSS between defense and alert}
\label{table:runtime}
\end{table*}

\noindent
\textbf{\large GraphSS Defends Node Classifiers.}
We examine how far GraphSS can defend node classifiers against dynamic graph perturbations. We randomly sample 20\% of the nodes from the test graph to build a subgraph and repeat this process five times to construct a group of dynamic subgraphs. For each subgraph, we apply non-targeted adversarial attacks on both links and features together ($L\&F$) \citep{zugner2018adversarial}. To assess the generalization, we apply the adversarial attacks on two groups of popular node classifiers, spectral models (GCN \citep{kipf2016semi}, SGC \citep{wu2019simplifying}) and spatial models (GraphSAGE \citep{hamilton2017inductive}), across five public datasets. We also compare the defending performance of GraphSS with the competing methods and present the accuracy of each group of dynamic subgraphs in Table \ref{table:exp1}. The best defending performance on each dataset is highlighted.
Originally, all three node classifiers have comparable performance. After the adversarial attacks, the results explicitly state that GraphSS gains superior defense against the state-of-the-art competing methods and even exceeds the original performance in some cases. The extent of recovery may sometimes be influenced by the attacking consequence, i.e., GraphSS can retrieve higher accuracy when the node classifier undergoes less degradation.

In this examination, we have several observations as follows.
1) AdvTrain gets unfavorable performance across all datasets, which reveals that simply using adversarial samples to train the node classifiers has limited contributions to the defense.
2) RGCN performs poorly in comparison to AdvTrain along with the density (Avg.D) of the graph increases because RGCN adopts a sampling in the hidden representation, whereas this sampling may lose more information on denser graphs.
3) GNN-SVD obtains lower accuracy than GNN-Jaccard across all graphs as GNN-SVD is tailored to fit the targeted attacks and may not adapt well to the non-targeted attacks.
4) GRAND deteriorates seriously on Cora. One of the reasons for this deterioration is that GRAND assumes the graph satisfies the homophily property. The assumption may break when the graph suffers serious perturbations. Besides, GRAND is sensitive to the parameter’s setting as we spent substantial time tuning the optimal parameters on different datasets. On the contrary, GraphSS does not assume such network property, and it is also easy to tune.
5) ProGNN is superior to GNN-Jaccard on denser graphs. We argue that ProGNN can simultaneously update the graph structure and the parameters of the node classifiers with the low rank and feature smoothness constraints. Such an update reaps huge fruits for recovering better graph structures on denser graphs.

\begin{figure}[h] 
  \centering
  \includegraphics[width=\linewidth]{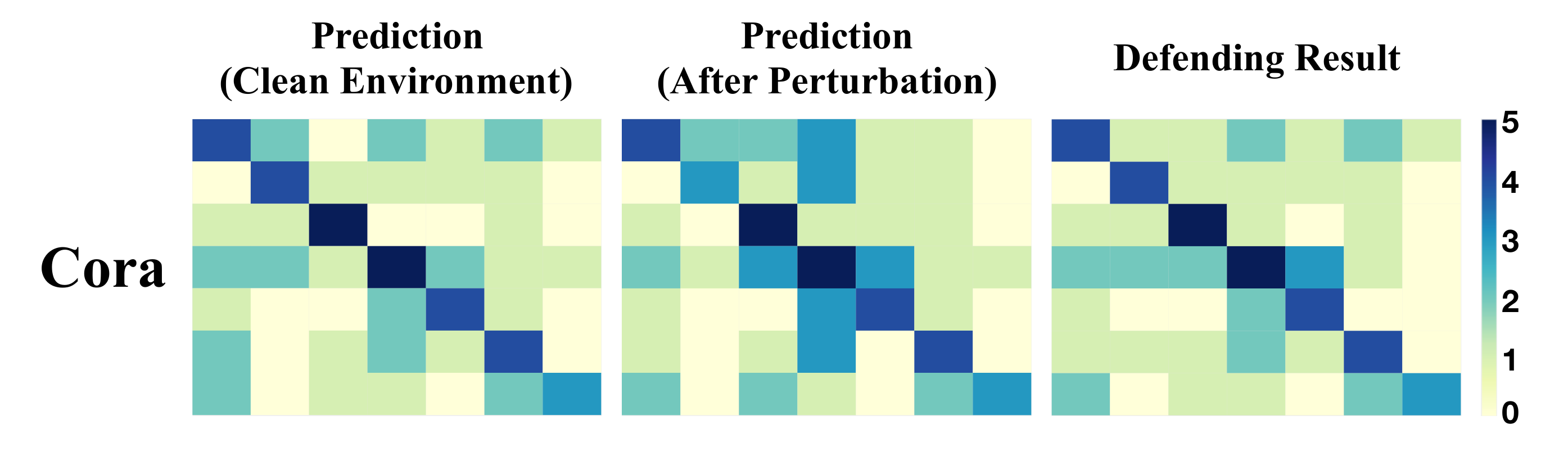}
  \caption{The confusion matrices (heatmap) of the defending result on Cora by GraphSS (We apply log-scale to the confusion matrix for fine-grained visualization.)}
\label{fig:fig_cm}
\end{figure}

Furthermore, we visualize our defending result (on top of GCN) on the test graph of Cora as an example in Figure \ref{fig:fig_cm}, which presents the result of prediction before/after graph perturbations, and our defending result. The visualization clarifies that GraphSS can recover the prediction after perturbations as close as that in a clean environment.

\begin{figure*}[t] 
  \hfill
  \begin{subfigure}{0.195\textwidth}
  \centering 
    \includegraphics[width=\linewidth]{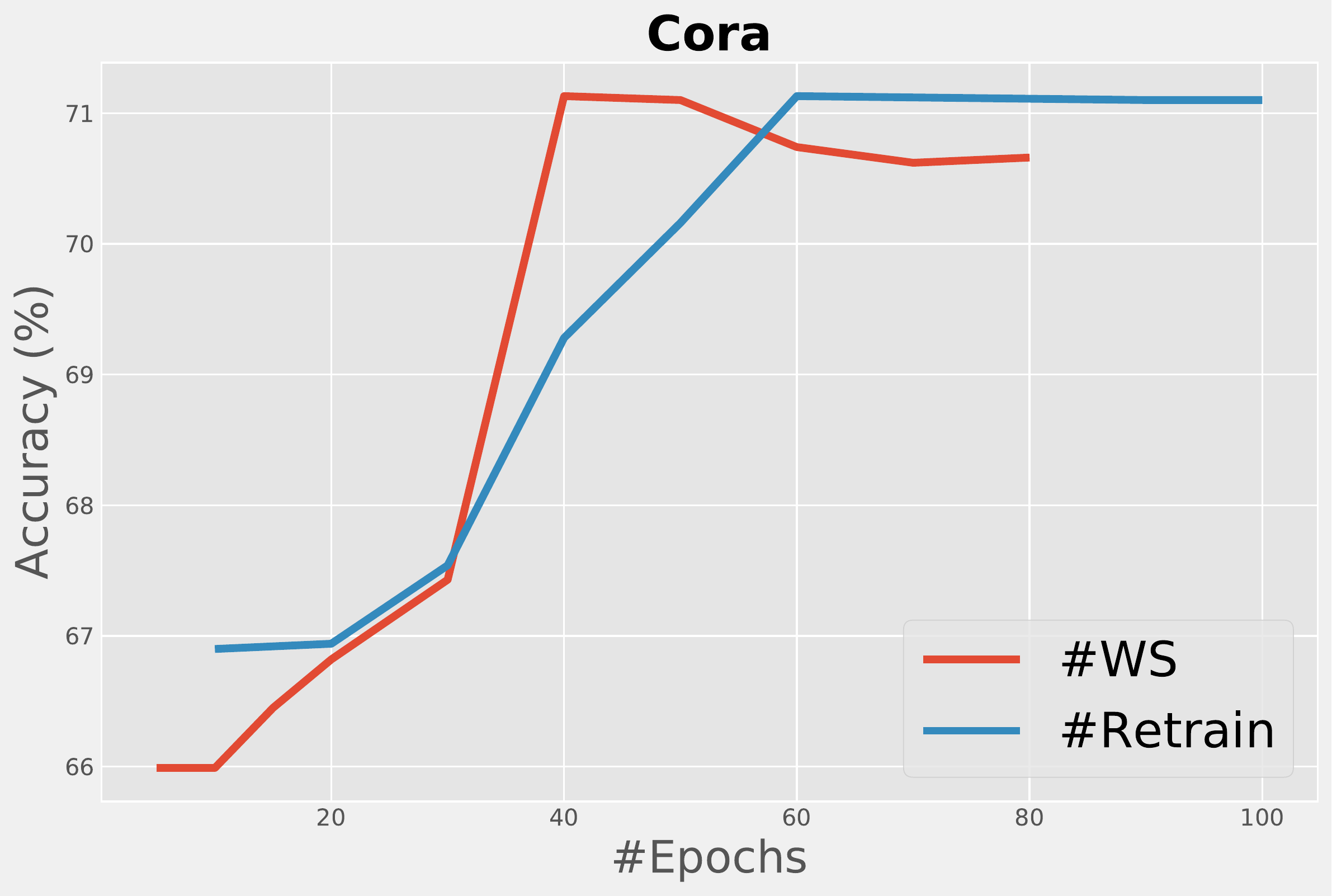}
  \end{subfigure}%
  \hfill
  \begin{subfigure}{0.195\textwidth}
  \centering 
    \includegraphics[width=\linewidth]{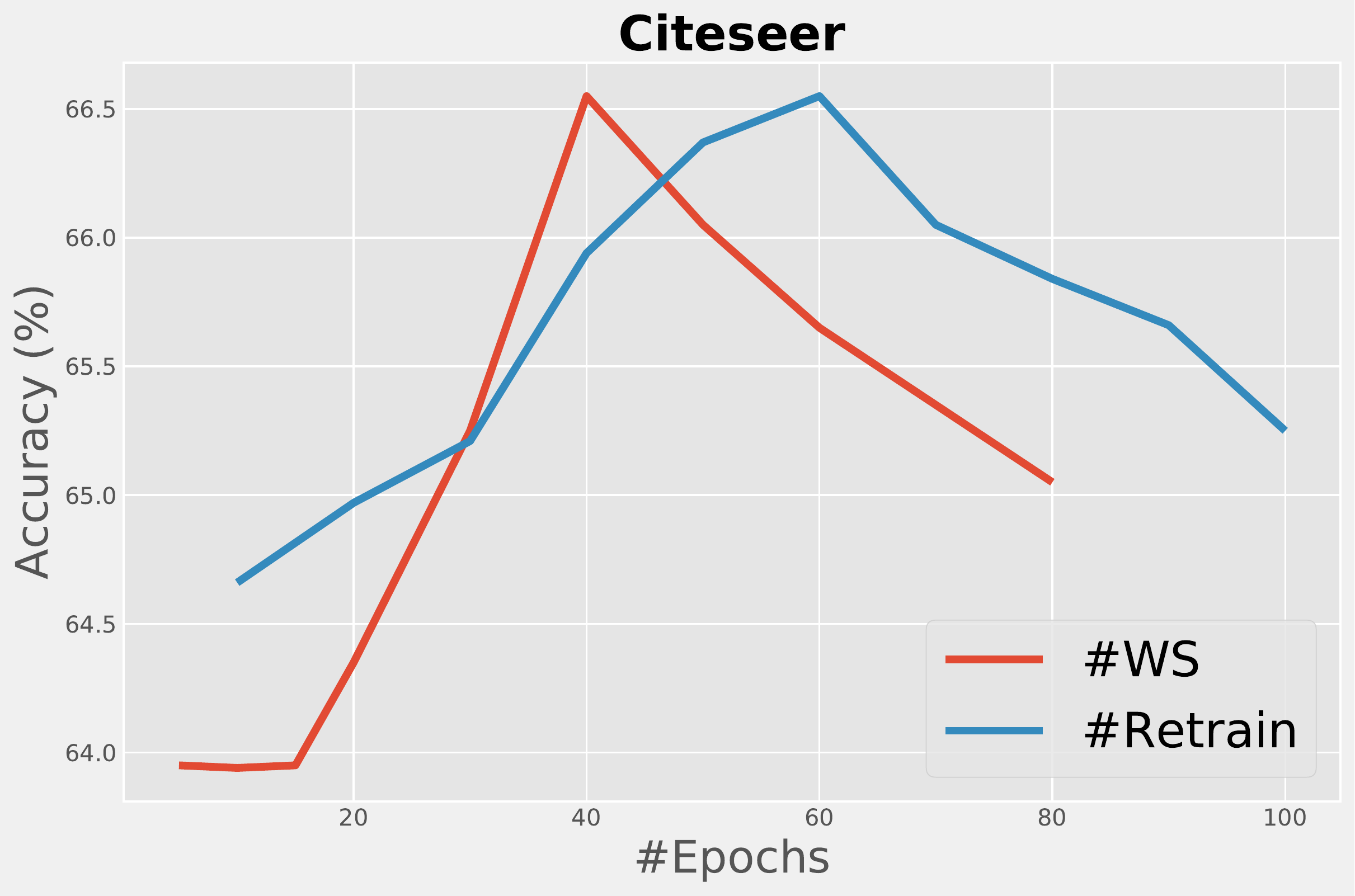}
  \end{subfigure}
  \begin{subfigure}{0.195\textwidth}
  \centering 
    \includegraphics[width=\linewidth]{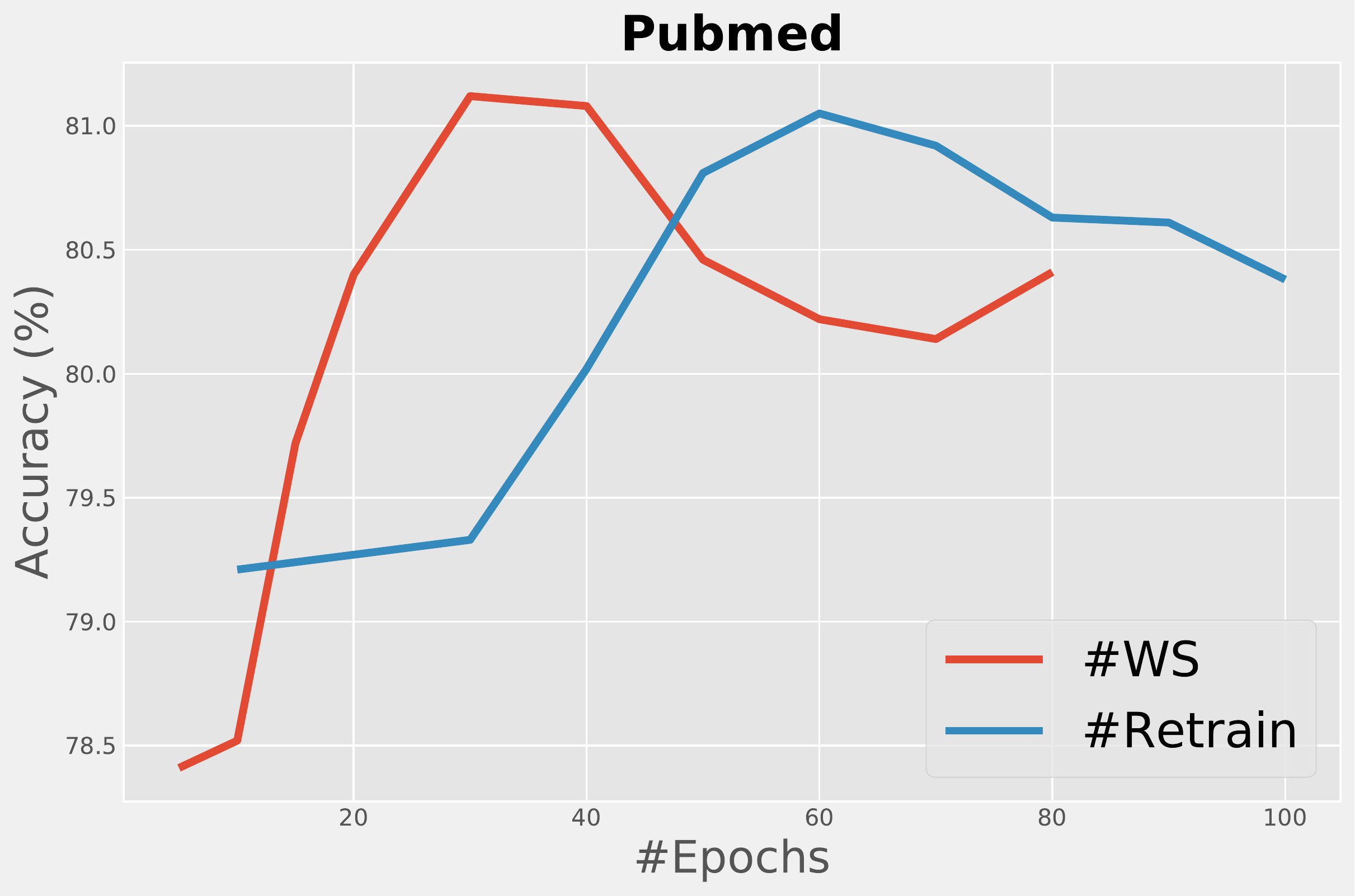}
  \end{subfigure}
  \begin{subfigure}{0.195\textwidth}
  \centering 
    \includegraphics[width=\linewidth]{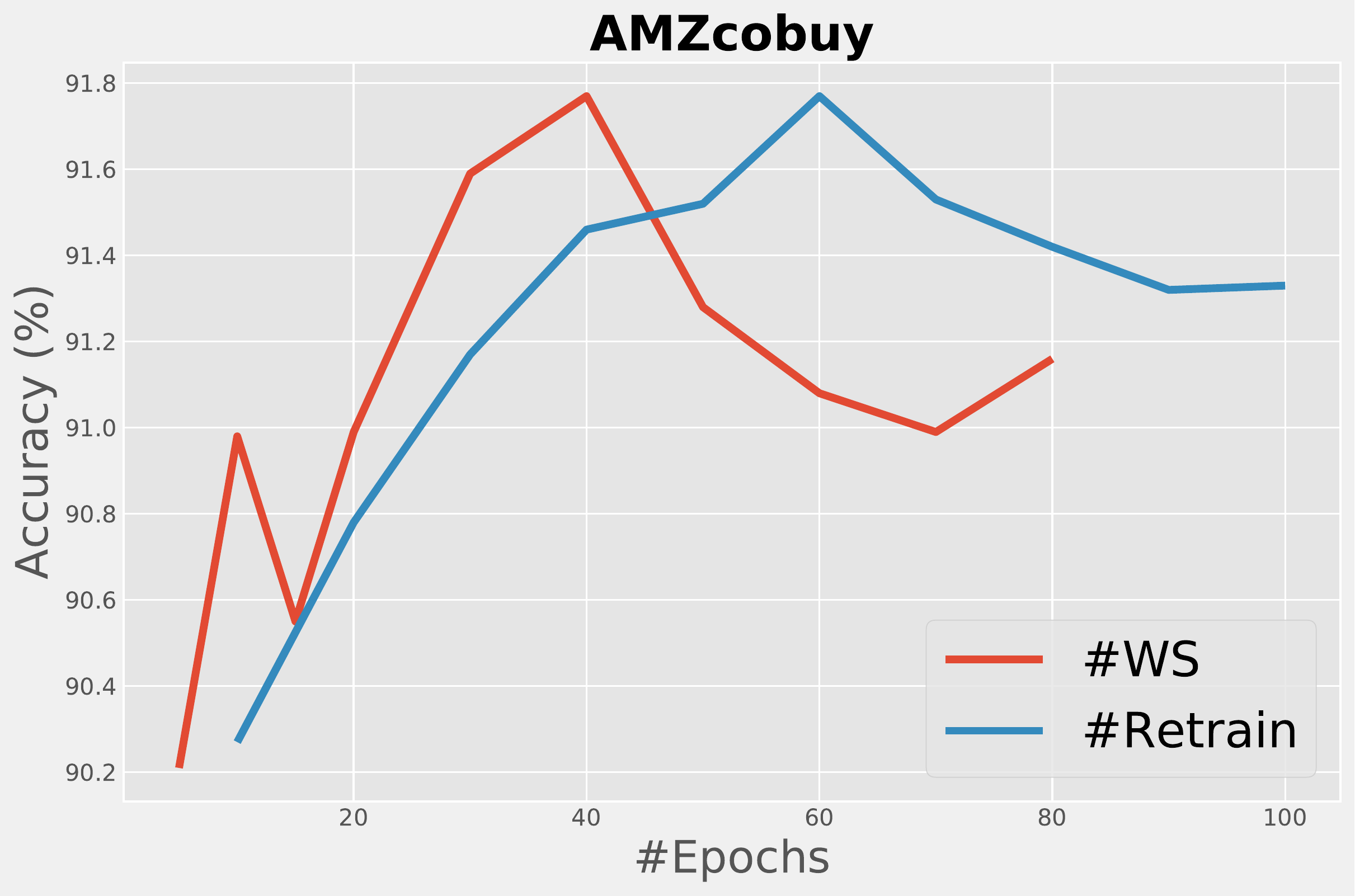}
  \end{subfigure}
  \begin{subfigure}{0.195\textwidth}
  \centering 
    \includegraphics[width=\linewidth]{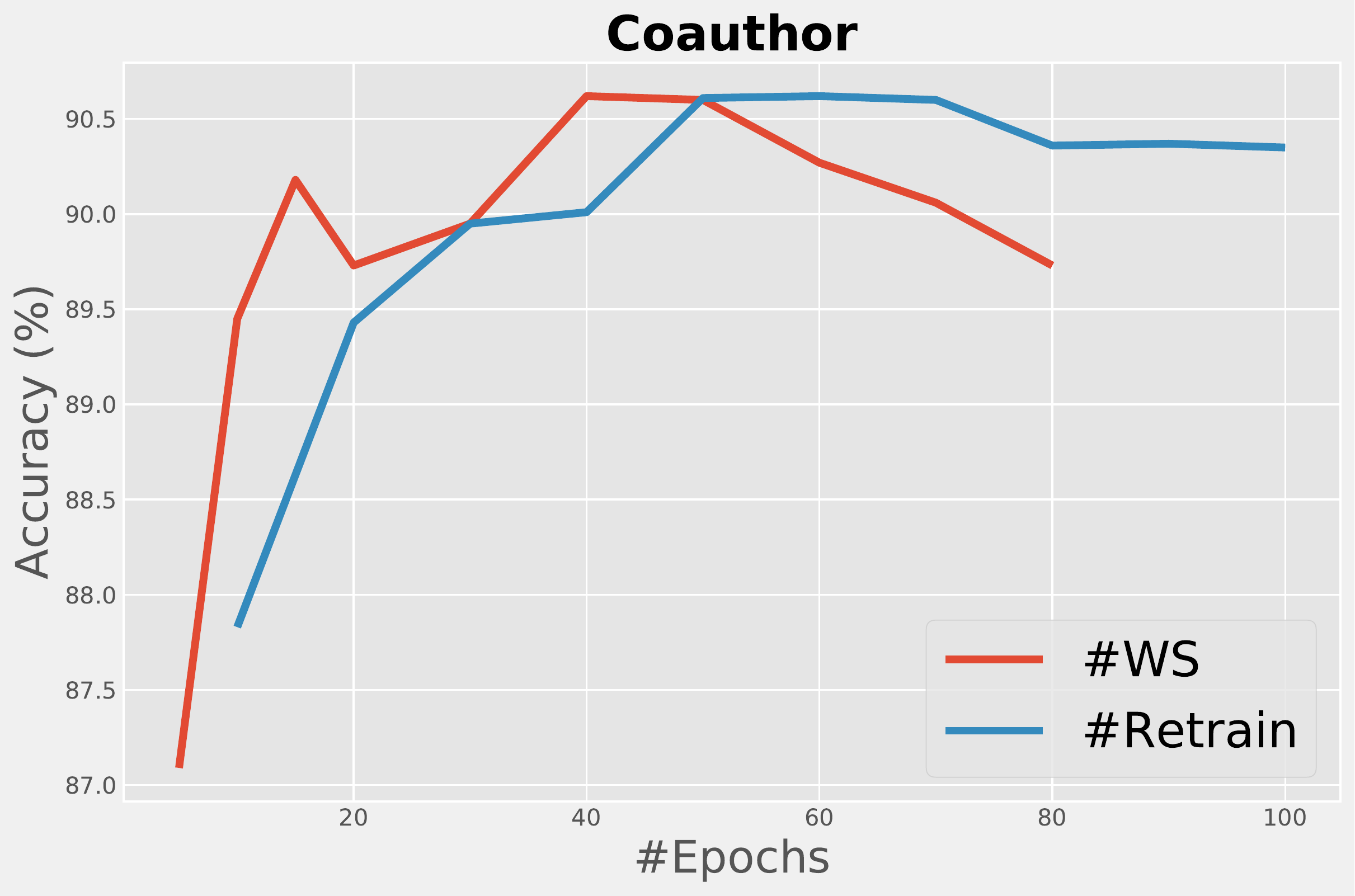}
  \end{subfigure}
\caption{Analysis of the number of warm-up steps $WS$ (blue) and the number of retraining epochs $Retrain$ (red) for GraphSS}
\label{fig:fig_hp}
\end{figure*}

\noindent
\textbf{\large GraphSS Can Alert Dynamic Graph Perturbations.}
We assess whether GraphSS can alert the perturbations on dynamic graphs. We repeat the aforementioned procedure ten times to construct a group of dynamic subgraphs and then randomly select three subgraphs from the group to be perturbed by non-targeted adversarial attacks ($L\&F$) \citep{zugner2018adversarial}. We assess our model on three node classifiers across five public datasets and present the performance via ROC curve and AUC score in Figure \ref{fig:fig_roc}. A higher true positive rate (TPR) indicates that GraphSS has a higher probability to successfully alert the true perturbations. On the contrary, a higher false positive rate (FPR) implies that GraphSS has more chance to make a false alert on unperturbed subgraphs. Figure \ref{fig:fig_roc} reveals that GraphSS achieves robust alerts on top of GCN across all datasets. For the other two node classifiers, GraphSS has higher TPR and AUC scores with GraphSAGE than with SGC. This indicates that GraphSS accomplishes better alert performance on top of GraphSAGE when the subgraph is undergoing perturbed.

\noindent
\textbf{\large Analysis of Runtime.}
We analyze the runtime of GraphSS between defense and alert. For each dataset in Table \ref{table:runtime}, the left column presents the average runtime whereas the right column presents the unit runtime. It's expected that GraphSS has longer runtime on a larger graph. However, the unit runtime maintains a comparable level as the size of the graph increases. This result indicates that the size of the graph doesn't affect the speed of inference. We observe that the average runtime of defense is slightly higher than that of alert. We argue that such a gap is reasonable as GraphSS iteratively retrains the node classifiers in the defense scenario. Overall, this analysis illustrates that GraphSS retains a stable speed of inference regardless of the size of the graph.
We also sort the defense runtime of 5 methods on Cora as an example in Table \ref{tab:model_runtime}. The sorting indicates that topological denoising methods mainly apply on the preprocessing stage and may have faster speed on smaller graphs.

\begin{table}[h]
\footnotesize
\centering
\setlength{\tabcolsep}{2pt}
\begin{tabular}{ccccc}
 \toprule
  \textbf{GNN-Jaccard} & \textbf{GNN-SVD} & \textbf{GraphSS} & \textbf{RGCN} & \textbf{GRAND} \\
  \midrule
    9.39{\scriptsize(±0.42)} & 10.72{\scriptsize(±0.55)} & 22.75{\scriptsize(±0.65)} & 27.65{\scriptsize(±1.36)} & 53.42{\scriptsize(±2.68)} \\
 \bottomrule
\end{tabular}
\caption{Runtime (s) comparison among 5 methods on Cora}
\label{tab:model_runtime}
\end{table}

\noindent
\textbf{\large Analysis of Parameters.}
We investigate how the number of warm-up steps $WS$ and the number of retraining epochs $Retrain$ affects the defending performance. We select GCN as the node classifier and conduct this analysis on the validation set. We fix the number of inference epochs as 100 and then apply grid search to look for the optimal parameters, where $WS \in [5, 80]$ and $Retrain \in [20, 100]$. The results turn out that $WS$, $Retrain$ reach the optimum nearby 40, and 60, respectively. To display the trend clearly, we fix one parameter as its optimal value and present another one in a curve. For example, we fix $Retrain$ as 60 and investigate how the validation accuracy changes with different $WS$ in the blue curve. To enlarge the difference, we display these curves separately in Figure \ref{fig:fig_hp}. We observe that both larger and smaller $WS$ have a negative effect on accuracy. Larger $WS$ means insufficient inference, whereas smaller $WS$ implies inadequate epochs to build the dynamic label transition matrix $\phi$. This property also applies to $Retrain$. Larger $Retrain$ is even harmful to the performance. In this study, we select the aforementioned optimal values for our model.

\begin{figure}[h] 
  \centering
  \includegraphics[width=\linewidth]{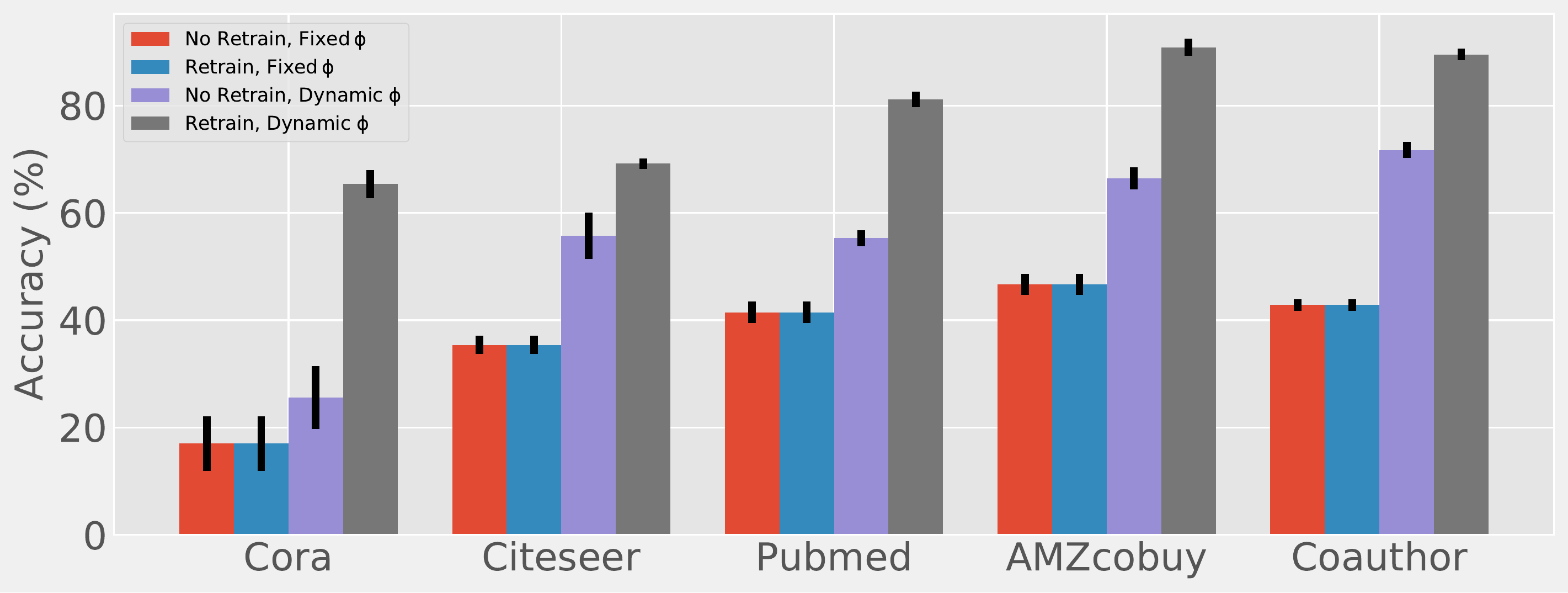}
  \caption{Ablation study of GraphSS (\{Retrain, No Retrain\} denote whether we retrain the node classifier in each iteration. \{Fixed $\phi$, Dynamic $\phi$\} denote whether we dynamically update the transition matrix $\phi$ in each iteration.)}
\label{fig:fig_bar}
\end{figure}

\noindent
\textbf{\large Ablation Study.}
To better understanding the model architecture, we conduct an ablation study on top of GCN to illustrate how the label transition matrix $\phi$ and retraining affect the defending performance of GraphSS under four scenarios. We follow the same procedure as our previous defense experiments. 
According to Figure \ref{fig:fig_bar}, we observe that the accuracy has no change without dynamically updating the transition matrix $\phi$. In the last two scenarios (Dynamic $\phi$), we notice that retraining the node classifier can help increase the accuracy further. This study reveals that dynamic transition matrix $\phi$ is a crucial component of GraphSS. Retraining can further promote the robustness of the node classifier with the help of dynamic $\phi$.

\noindent
\textbf{\large Limitation and Future Directions.}
1) The inference of GraphSS depends on the warm-up label transition matrix $\phi'$, which is built with the categorical distribution on the train graph. Such dependence indicates that GraphSS can only handle the evasion attacks at this point. In the future, GraphSS can be improved to defend the poisoning attacks by iteratively updating the auto-generated labels. Besides, this update may benefit the node classifier under the label scarcity scenario for self-training purposes.
2) On the other hand, GraphSS can only alert the perturbations in this version. However, such a version could be extended to detect the anomaly nodes on dynamic graphs.

\section{Related Work}
\label{sec:rewk}

Related works about the defense of GCNs are already discussed in {\bf Introduction}. Due to the page limit, we only present the studies about learning deep neural networks with noisy labels in this section \cite{reed2014training, goldberger2016training, patrini2017making, yao2019safeguarded, sukhbaatar2014training, misra2016seeing}. Those works achieve great improvement in conventional CNNs.

\section{Conclusion}
\label{sec:con}
In this paper, we generalize noisy supervision as a subset of self-supervised learning methods. This generalization regards the noisy labels, including both manual-annotated labels and auto-generated labels, as one kind of self-information for each node. The robustness of the node classifier can be improved by utilizing such self-information.
We then propose a new Bayesian self-supervision model, namely GraphSS, to accomplish this improvement by supervising the categorical distribution of the latent labels based on dynamic conditional label transition, which follows the Dirichlet distribution.
Extensive experiments demonstrate that GraphSS can not only affirmatively alert the perturbations on dynamic graphs but also effectively recover the prediction of a node classifier when the graph is under such perturbations. These two advantages prove to be generalized over three classic GCNs across five public graph datasets.

\section{Acknowledgments}
This research is partially supported by National Science Foundation with grant number IIS-1909916.


\section{Community Impact}
Many existing defending methods confront the adversarial perturbations from the perspective of topological denoising and message-passing mechanism. The low-hanging fruits from these two aspects have been widely explored. On the contrary, our model improves the robustness of the node classifiers by recovering the node distribution. 
Also, incorporating self-supervision into the defense of GCNs is a promising direction in the recent two years. Our work provides an innovative alternative in this path and could be a baseline model for the above-mentioned future directions.

\appendix
\section{Appendix}
\label{sec:app}

\subsection{Hardware and Software}
All above-mentioned experiments are conducted on the server with the following configurations:
\begin{itemize}
  \item Operating System: Ubuntu 18.04.5 LTS
  \item CPU: Intel(R) Xeon(R) Gold 6258R CPU @ 2.70 GHz
  \item GPU: NVIDIA Tesla V100 PCIe 16GB
  \item Software: Python 3.8, PyTorch 1.7.
\end{itemize}

\subsection{Implementation of Dynamic Graph Perturbations}
To simulate the dynamic graph perturbations, we execute node-level direct evasion non-targeted attacks \citep{zugner2018adversarial} on the links and features of the nodes ($L\&F$) in dynamic graphs.
Similar to \citep{you2020does}, we denote the intensity of perturbation as $n_{pert}$. We set $n_{pert}$ for Cora, Citeseer, and Pubmed as 2, and for AMZcobuy, Coauthor as 5. The ratio of $n_{pert}$ between applying on links and applying on features is 1 : 10. For example, the attacker applies 2 perturbations on links and 20 perturbations on features for $L\&F$. To construct a group of dynamic subgraphs, we select the random seed from 1 to $n_{graphs}$, where $n_{graphs}$ denotes the number of dynamic subgraphs.

\subsection{Model Architecture}
Our proposed model, GraphSS, can be applied on top of GCNs. The model architecture and hyper-parameters of GCNs are described in Table \ref{tab:gcn_archit} and Table \ref{tab:hp}, respectively. The hyper-parameters of GraphSS, $\alpha$, is fixed as 1.0.

\begin{table}[h]
\centering
\begin{tabular}{cc}
  \toprule
    \textbf{Hyper-parameters} & \textbf{Values} \\
    \midrule
    \textbf{\#Layers} & 2 \\
    \textbf{\#Hidden} & 200 \\
    \textbf{Optimizer} & Adam \\
    \textbf{Learning Rate} & $1 \times 10^{-3}$ \\
  \bottomrule
\end{tabular}
\caption{Hyper-parameters of GCNs (\#Hidden denotes the number of neurons in each hidden layer of GCNs)}
\label{tab:hp}
\end{table}

\begin{table}[h]
\footnotesize
\centering
\setlength{\tabcolsep}{4pt}
\begin{tabular}{ccccc}
 \toprule
  \textbf{Model} & \textbf{Aggregator} & \textbf{\#Hops} & \textbf{Activation} & \textbf{Dropout} \\
  \midrule
    $\mathbf{GCN}$ & $\times$ & $\times$ & ReLU & 0.0 \\
    $\mathbf{SGC}$ & $\times$ & 2 & $\times$ & 0.0 \\
    $\mathbf{GraphSAGE}$ & gcn & $\times$ & ReLU & 0.0 \\
 \bottomrule
\end{tabular}
\caption{Model architecture of GCNs}
\label{tab:gcn_archit}
\end{table} 

\begin{table}[h]
\scriptsize
\centering
\setlength{\tabcolsep}{2pt}
\begin{tabular}{cccccc}
 \toprule
  \textbf{Hyper-parameters} & \textbf{Cora} & \textbf{Citeseer} & \textbf{Pubmed} & \textbf{AMZcobuy} & \textbf{Coauthor} \\
 \midrule
    DropNode probability & 0.5 & 0.5 & 0.5 & 0.5 & 0.5 \\
    Propagation step & 8 & 2 & 5 & 5 & 5 \\
    Data augmentation times & 4 & 2 & 4 & 3 & 3 \\
    CR loss coefficient & 1.0 & 0.7 & 1.0 & 0.9 & 0.9 \\
    Sharpening temperature & 0.5 & 0.3 & 0.2 & 0.4 & 0.4 \\ 
    Learning rate & 0.01 & 0.01 & 0.2 & 0.2 & 0.2 \\
    Early stopping patience & 200 & 200 & 100 & 100 & 100 \\
    Hidden layer size & 32 & 32 & 32 & 32 & 32 \\
    L2 weight decay rate & 5e-4 & 5e-4 & 5e-4 & 5e-4 & 5e-4 \\
    Dropout rate in input layer & 0.5 & 0.0 & 0.6 & 0.6 & 0.6 \\
    Dropout rate in hidden layer & 0.5 & 0.2 & 0.8 & 0.5 & 0.5 \\
 \bottomrule
\end{tabular}
\caption{Hyper-parameters of GRAND in this paper}
\label{tab:grand_para}
\end{table} 

\subsection{Hyper-parameters of Competing Methods}
We present the hyper-parameters of our competing methods below. For reproducibility, we maintain the same denotation for each competing method as the corresponding original paper. The competing models are trained by Adam optimizer with 200 epochs.
\begin{description}
\item[$\bullet$\ AdvTrain~\cite{wang2019graphdefense, you2020does}] assigns pseudo labels to generated adversarial samples and then retrains the node classifier with both noisy labeled nodes and adversarial nodes. In detail, we first generate the adversarial samples by $L\&F$. The number of adversarial samples is equal to 10\% of victim nodes. After that, we add up these generated adversarial samples into the training set and then retrain the node classifier. Finally, we evaluate this defending result by implementing $L\&F$ on the victim nodes. The approach of adversarial training can be formulated as follows:
\begin{equation}
\begin{aligned}
\mathcal{L}_{noisy} = \mathcal{L}(\mathbf{Y}_{noisy}, \  \textit{f}_{\theta}(\mathbf{\tilde{A}}, \mathbf{X})), \\
\mathcal{L}_{adv} = \mathcal{L}(\mathbf{Y}_{pseudo}, \  \textit{f}_{\theta}(\mathbf{A}', \mathbf{X}')), \\
\theta^{*} = \arg \min_{\theta} \left( \mathcal{L}_{noisy} + \mathcal{L}_{adv} \right),
\label{eqn:loss_adv}
\end{aligned}
\end{equation}
where $\mathbf{Y}_{noisy}$, $\mathbf{Y}_{pseudo}$ are the corresponding labels to compute loss functions, and $\mathbf{A}'$, $\mathbf{X}'$ are perturbed adjacency, feature matrices generated by the attack algorithm.
\item[$\bullet$\ GNN-Jaccard~\cite{wu2019adversarial}] preprocesses the graph by eliminating suspicious connections, whose Jaccard similarity of node’s features is smaller than a given threshold. The similarity threshold for dropping edges is 0.01. \#Hidden is 16. The dropout rate is 0.5.
\item[$\bullet$\ GNN-SVD~\cite{entezari2020all}] proposes another preprocessing approach with low-rank approximation on the perturbed graph to mitigate the negative effects from high-rank attacks, such as Nettack~\citep{zugner2018adversarial}. The number of singular values and vectors is 15. \#Hidden is 16. The dropout rate is 0.5.
\item[$\bullet$\ RGCN~\cite{zhu2019robust}] adopts Gaussian distributions as the hidden representations of nodes to mitigate the negative effects of adversarial attacks. We set up $\gamma$ as 1, $\beta_1$ and $\beta_2$ as $5 \times 10^{-4}$ on all datasets. \#Hidden is 32. The dropout rate is 0.6. The learning rate is 0.01.
\item[$\bullet$\ GRAND~\cite{feng2020graph}] proposes random propagation and consistency regularization strategies to address the issues about over-smoothing and non-robustness of GCNs. We follow the same procedure to tune the hyper-parameters. The optimal hyper-parameters of GRAND in this paper are reported in Table~\ref{tab:grand_para}.
\item[$\bullet$\ ProGNN~\cite{jin2020graph}] jointly learns the structural graph properties and iteratively reconstructs the clean graph to reduce the effects of adversarial structure. We select $\alpha$, $\beta$, $\gamma$, and $\lambda$ as $5 \times 10^{-4}$, 1.5, 1.0, and $1 \times 10^{-3}$, respectively. \#Hidden is 16. The dropout rate is 0.5. The learning rate is 0.01. Weight decay is $5 \times 10^{-4}$.
\item[$\bullet$\ NRGNN~\cite{dai2021nrgnn}] develops a label noise-resistant framework that brings clean label information to unlabeled nodes based on the feature similarity. We follow the corresponding settings except that we choose uniform noise with ptb\_rate=0.1.
\end{description}

\bibliography{2reference}

\end{document}